\pdfoutput=1

\documentclass[11pt]{article}

\usepackage{ACL2023}
\usepackage{times}
\usepackage{latexsym}

\usepackage[T1]{fontenc}

\usepackage[utf8]{inputenc}

\usepackage{microtype}

\usepackage{inconsolata}

\usepackage{booktabs,microtype,xspace,graphicx,amsmath,array,multirow,makecell,enumitem,xparse}
\usepackage{caption}
\usepackage{subcaption}
\usepackage[toc,page]{appendix}
\usepackage{framed}
\usepackage{soul}
\usepackage[utf8]{inputenc}
\usepackage{pifont}

\NewDocumentCommand{\rot}{O{45} O{1em} m}{\makebox[#2][l]{\rotatebox{#1}{#3}}}%
\usepackage{numprint}
\npthousandsep{,}

\newcommand{\emldisplay}[2]{\texttt{\href{mailto:#1}{#2}}}
\newcommand{\eml}[1]{\emldisplay{#1}{#1}}
\newcommand{\finalversion}[1]{\unskip}

\makeatletter
\renewcommand{\paragraph}{%
  \@startsection{paragraph}{4}{\z@}{0.5ex plus%
   0.5ex minus .2ex}{-1em}{\normalsize\bf}%
}
\makeatother

\newcommand{\pretrained}[0]{pre-trained\xspace}

\newcommand{\pretraining}[0]{pre-training\xspace}
\newcommand{\nonpretrained}[0]{non-pretrained\xspace}

\newcommand{\humanconstructedbenchmarks}[0]{\humanconstructed{} benchmarks\xspace}
\newcommand{\humanconstructed}[0]{human-constructed\xspace}

\newcommand{\Humanconstructedbenchmarks}[0]{Human-constructed benchmarks\xspace}
\newcommand{\HumanConstructedBenchmarks}[0]{Human-Constructed Benchmarks\xspace}
\newcommand{\approach}[0]{approach\xspace}
\newcommand{\approaches}[0]{\approach{}es\xspace}
\newcommand{\Approach}[0]{Approach\xspace}
\newcommand{\Approaches}[0]{\Approach{}es\xspace}
\newcommand{\modelingapproach}[0]{modeling \approach{}\xspace}
\newcommand{\modelingapproaches}[0]{modeling \approaches{}\xspace}
\newcommand{\ModelingApproaches}[0]{Modeling \Approaches{}\xspace}
\newcommand{\numModels}[0]{20\xspace}

\newcommand{\numHyperparameters}[0]{5\xspace}
\newcommand{\wikidata}[0]{Wikidata\xspace}
\newcommand{\wikidatasyntheticqa}[0]{WikidataSyntheticQA\xspace}
\newcommand{\fuzzypmsyntheticqa}[0]{FuzzySyntheticQA\xspace}
\newcommand{\numBenchmarks}[0]{32\xspace}
\newcommand{\Dtrain}{\ensuremath{D_\text{train}}}
\newcommand{\Dtest}{\ensuremath{D_\text{test}}}
\newcommand{\A}{\ensuremath{\mathcal{A}}}
\newcommand{\PreserveBackslash}[1]{\let\temp=\\#1\let\\=\temp}
\newcolumntype{C}[1]{>{\PreserveBackslash\centering}p{#1}}
\newcolumntype{R}[1]{>{\PreserveBackslash\raggedleft}p{#1}}
\newcolumntype{L}[1]{>{\PreserveBackslash\raggedright}p{#1}}

\definecolor{lightgray}{gray}{0.9}
\definecolor{passage}{HTML}{FFC7BF}
\definecolor{answer}{HTML}{CFFFCC}
\definecolor{question}{HTML}{E3C0FF}
\definecolor{question1}{HTML}{B6D7A8}
\definecolor{question2}{HTML}{FFE599}
\definecolor{question3}{HTML}{D5A6BD}
\newcommand{\hlc}[2][yellow]{{%
    \colorlet{foo}{#1}%
    \sethlcolor{foo}\hl{#2}}%
}
\newcommand\independent{\protect\mathpalette{\protect\independenT}{\perp}}
\def\independenT#1#2{\mathrel{\rlap{$#1#2$}\mkern2mu{#1#2}}}
\newcommand{\cmark}{\ding{51}}
\newcommand{\xmark}{\ding{55}}

\newif\ifcomments
\commentstrue

\ifcomments
    \providecommand{\nfl}[1]{{\protect\color{Green}{[NFL: #1]}}}
    \providecommand{\tl}[1]{{\protect\color{Magenta}{[TL: #1]}}}
    \providecommand{\rj}[1]{{\protect\color{ProcessBlue}{[RJ: #1]}}}
    \providecommand{\pl}[1]{{\protect\color{Red}{[PL: #1]}}}
\else
    \providecommand{\nfl}[1]{}
    \providecommand{\tl}[1]{}
    \providecommand{\rj}[1]{}
    \providecommand{\pl}[1]{}
\fi

\title{Do Question Answering Modeling Improvements\\Hold Across Benchmarks?}

\author{
    Nelson F. Liu$^{\spadesuit}$ \quad
    Tony Lee$^{\spadesuit}$ \quad
    {\bf Robin Jia}$^{\heartsuit}$ \quad
	{\bf Percy Liang}$^{\spadesuit}$\\
    $^\spadesuit$Computer Science Department, Stanford University, Stanford, CA\\
	$^\heartsuit$Department of Computer Science, University of Southern California, Los Angeles, CA\\
	\{\emldisplay{nfliu@cs.stanford.edu}{nfliu}, \emldisplay{tonyhlee@cs.stanford.edu}{tonyhlee}, \emldisplay{pliang@cs.stanford.edu}{pliang}\}\texttt{@cs.stanford.edu}\\
  \eml{robinjia@usc.edu}}

\begin{document}
\maketitle
\begin{abstract}
  Do question answering (QA) modeling improvements (e.g., choice of architecture and training procedure) hold consistently across the diverse landscape of QA benchmarks?
  To study this question, we introduce the notion of \emph{concurrence}---two benchmarks have high concurrence on a set of modeling approaches if they rank the \modelingapproaches{} similarly.
  We measure the concurrence between \numBenchmarks{} QA benchmarks on a set of \numModels{} diverse \modelingapproaches{} and find that human-constructed benchmarks have high concurrence amongst themselves, even if their passage and question distributions are very different.
  Surprisingly, even downsampled human-constructed benchmarks (i.e., collecting less data) and programmatically-generated benchmarks (e.g., cloze-formatted examples) have high concurrence with human-constructed benchmarks.
  These results indicate that, despite years of intense community focus on a small number of benchmarks, the modeling improvements studied hold broadly.
\end{abstract}

\section{Introduction}

\begin{figure}[t]
  \centering
  \includegraphics[width=\columnwidth]{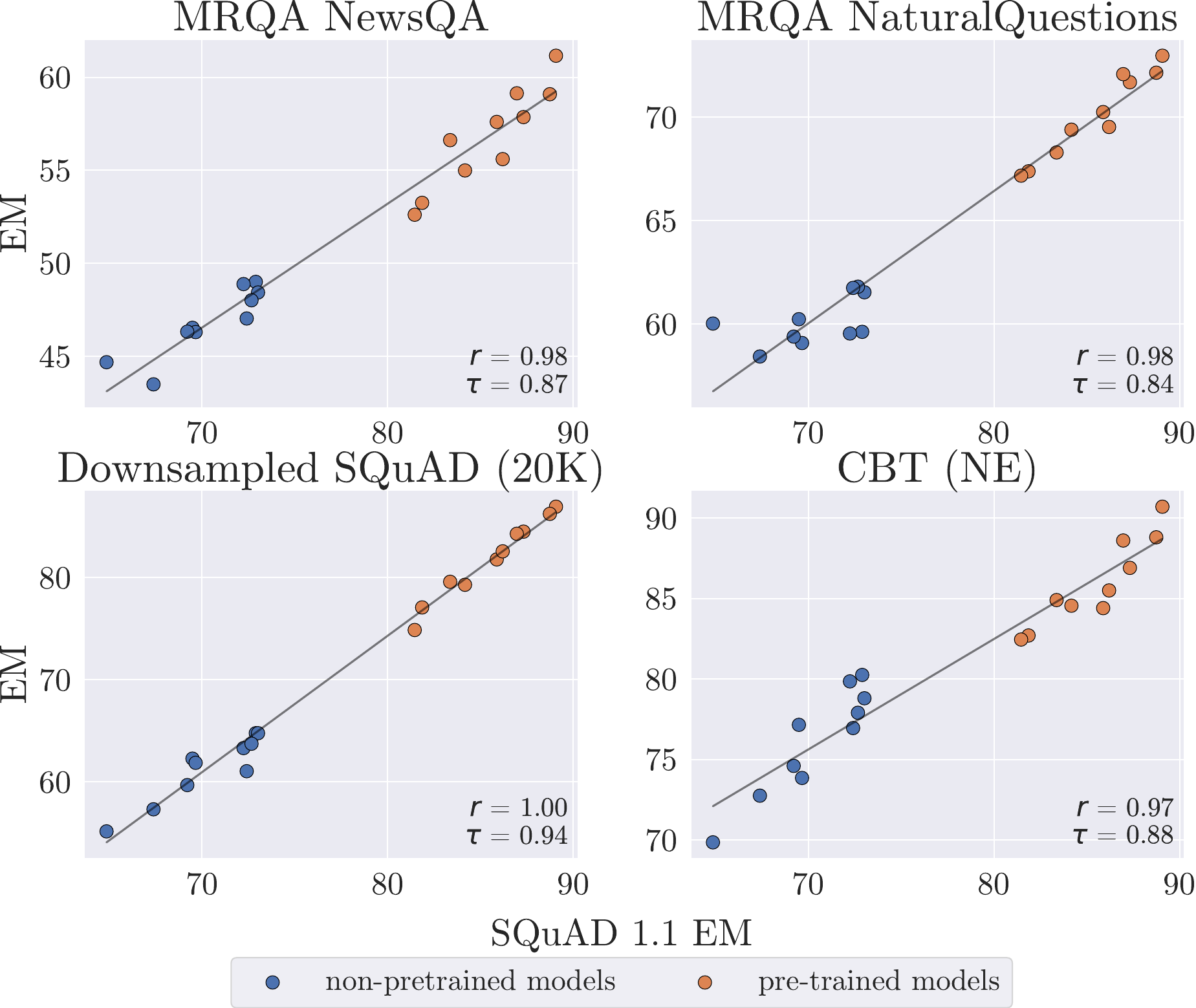}
  \caption{Two benchmarks have high concurrence if they rank a set of \modelingapproaches{} similarly. Surprisingly, we find that \humanconstructedbenchmarks{} (e.g., SQuAD, NaturalQuestions) have high concurrence with other \humanconstructedbenchmarks{}, downsampled \humanconstructedbenchmarks{}, and even programmatically-generated cloze benchmarks (e.g., the Children's Book Test; CBT). In addition, we are able to construct synthetic benchmarks that have high concurrence with \humanconstructedbenchmarks{} despite lacking natural language passages or questions.}
  \label{fig:results_summary}
\end{figure}

The NLP community has created a diverse landscape of extractive question answering (QA) benchmarks---their context passages may come from different sources, their questions may focus on different phenomena or be written by different populations, or other aspects of the data collection process may differ.
Driven to improve benchmark performance, researchers have proposed a variety of QA \modelingapproaches{}.
However, not all benchmarks receive equal attention from the community \citep{koch2021reduced}; many QA modeling approaches are developed on a small handful of benchmarks, especially those with popular leaderboards (e.g., SQuAD; \citealp{Rajpurkar2016SQuAD10}).
As a result, it is conceivable that some modeling improvements may not hold because they are (perhaps inadvertently) benchmark-specific, while others (e.g., pre-training on more data) hold more broadly.

In this work, we evaluate whether improvements from modeling \emph{approaches} hold (e.g., choices in model architecture or training procedure)---if a particular \modelingapproach{} improves performance when trained and evaluated on one benchmark, does it also improve performance on others?
Although much existing work studies whether \emph{systems} generalize (i.e., a model with a particular set of parameters;  \citealp{jia2017adversarial,talmor2019multiqa,miller2020effect}), research value often comes not from the systems themselves (e.g., model weights), but from the underlying ideas, techniques, and approaches.
We study the comparatively under-investigated question of whether such modeling \emph{approaches} generalize.

To study whether modeling improvements hold across benchmarks, we introduce the notion of \emph{concurrence}.
We say that two benchmarks have high concurrence on a set of modeling approaches if they rank the \modelingapproaches{} similarly.
To assess whether modeling improvements hold across the space of QA benchmarks, we measure the concurrence between \numBenchmarks{} diverse QA benchmarks on a testbed of \numModels{} representative \modelingapproaches{} introduced between 2016 and 2020.

Overall, we find that benchmarks that differ substantially still often have high concurrence.
Human-constructed benchmarks (e.g., SQuAD and MRQA NaturalQuestions) have high concurrence with each other, despite differences in crowdsourcing setups, passage and question distributions, and even linguistic phenomena of focus (\S\ref{sec:squad_vs_human_constructed}). 

How different can a benchmark be, while still maintaining high concurrence with \humanconstructedbenchmarks{}?
In \S\ref{sec:squad_vs_subsampled_squad}, we investigate the role of training dataset size by measuring concurrence with downsampled training datasets (e.g., using 20K SQuAD training examples rather than the full 88K).
We find that downsampled training datasets are sufficient for high concurrence with other human-constructed benchmarks. In \S\ref{sec:squad_vs_cloze}, we measure concurrence between human-constructed and programmatically-generated benchmarks (e.g., cloze-formatted or synthetic) to better understand the importance of human-written questions and passages.
We find that cloze-formatted benchmarks have high concurrence with \humanconstructedbenchmarks{}, so human-written questions and passages are not strictly necessary for concurrence. However, programmatically-generated synthetic benchmarks (e.g., the bAbI task suite) have low concurrence.
Having found this breaking point of low concurrence, we construct two minimal synthetic benchmarks that achieve high concurrence with \humanconstructedbenchmarks{}, despite lacking linguistic structure. Intuitively, the benchmarks that concur with \humanconstructedbenchmarks{} are those that require model capabilities that are also useful for better performance on \humanconstructedbenchmarks{} (e.g., identifying paraphrase and lexical overlap; \S\ref{sec:babi}-\ref{sec:squad_vs_wikidata}).

Our results have several implications for the future development of benchmarks and \modelingapproaches{}. To summarize:

\begin{enumerate}[itemsep=0em]
    \item Human-constructed benchmarks have high concurrence with each other on our testbed of \numModels{} \modelingapproaches{}. The \modelingapproaches{} studied are not particularly benchmark-specific and that their modeling improvements largely hold across different benchmarks, despite intense community focus on a small number of benchmarks. This is especially true of recent modeling improvements driven by better pre-training, which is largely downstream benchmark-agnostic.
    
    \item Many benchmarks require reasoning over predicate-argument structure (e.g., SQuAD, NewsQA, NaturalQuestions), and improvements on these benchmarks also transfer to more specialized benchmarks (e.g., HotpotQA or MRQA DROP) because (1)~almost all benchmarks involve reasoning over predicate-argument structure and/or (2)~better reasoning over predicate-argument structure is correlated with improvements on other phenomena.

    \item Human-constructed benchmarks are not strictly necessary for improving performance on other \humanconstructedbenchmarks{}. Synthetic benchmarks may be useful tools for isolating, understanding, and improving on particular model capabilities.
    
    \item Downsampling benchmarks to as few as 10K training examples does not significantly affect concurrence, especially since recent pre-trained modeling approaches have greater sample efficiency. We recommend the community build benchmarks that are smaller but more challenging (e.g., harder/more expensive to label per-example).
    
    \item Since \humanconstructedbenchmarks{} have high concurrence amongst themselves, we encourage researchers to seek diversity and build benchmarks that explore qualitatively different modeling capabilities that push research in new directions.
\end{enumerate}

\section{Measuring Concurrence}

Informally, we say that two benchmarks have high \emph{concurrence} on a set of \modelingapproaches{} if the two benchmarks rank the \modelingapproaches{} similarly.
We compare the performance of a \modelingapproach{} when trained and tested on one benchmark with its performance when trained and tested on another benchmark---we use each benchmark's original \emph{i.i.d.} train-test split, so all evaluation is in-domain. Repeating this process for many \modelingapproaches{}, we can assess whether performance gains \emph{between} modeling approaches are generally preserved when moving between benchmarks.

Formally, define a benchmark $B$ as a pair of datasets $(\Dtrain, \Dtest)$, where $\Dtrain \subseteq \mathcal{X} \times \mathcal{Y}$ and $\Dtest \subseteq \mathcal{X} \times \mathcal{Y}$ for an input space $\mathcal{X}$ and an output space $\mathcal{Y}$.
A \emph{system} is a function $s: \mathcal{X} \rightarrow \mathcal{Y}$ (i.e., a trained model with a particular set of parameters).
In contrast, a \emph{\modelingapproach{}} (i.e., a neural architecture coupled with a training procedure) is a function $a$ that takes in a training dataset $\Dtrain$ and outputs a system.
Let $\textsc{eval}$ denote an evaluation function, where $\textsc{eval}(a, B)$ returns the performance (under a given evaluation function, e.g., exact match) of a \modelingapproach{} $a$ when trained on the train split of $B$ and tested on the test split of $B$.
Finally, $\textsc{Concur}(B_{1}, B_{2}; \mathcal{A}, \textsc{eval})$ is the \emph{concurrence} between the benchmarks $B_{1}$ and $B_{2}$ with respect to a set of \modelingapproaches{} $\mathcal{A}$ and the evaluation function $\textsc{eval}$.
Let $a \sim \text{uniform}(\mathcal{A})$, where $\text{uniform}(\mathcal{A})$ denotes the uniform distribution over the set of \modelingapproaches{} $\A$.
Defining the random variables $P_{1} = \textsc{eval}(a, B_{1})$ and $P_{2} = \textsc{eval}(a, B_{2})$, we finally define
\begin{displaymath}
  \textsc{Concur}(B_{1}, B_{2}; \A, \textsc{eval}) = \textsc{corr}(P_{1}, P_{2})\,,
\end{displaymath}
where \textsc{corr} is some correlation function.

We use the SQuAD exact match (EM) metric as our evaluation function \textsc{eval}, and we consider the Pearson correlation coefficient ($r$) and the Kendall rank correlation coefficient ($\tau$) as our correlation functions \textsc{corr}.
The former measures whether the relationship between model performance on the two benchmarks is approximately linear, whereas the latter measures whether pairwise rank comparisons between models are preserved between benchmarks. As a rough guideline, we consider $\tau > 0.8$ to be high concurrence, though interpreting concurrence often requires more than comparing overall correlation.

\paragraph{Extractive QA modeling approaches.} To assess concurrence in this work, we use a representative set of \numModels{} diverse \modelingapproaches{} introduced between 2016 to 2020 (\A). These \modelingapproaches{} include RaSoR \citep{lee2016learning}, BiDAF \citep{seo2017bidirectional}, DocumentReader \citep{chen2017reading}, QANet \citep{yu2018qanet}, BiDAF++ \citep{clark-gardner-2018-simple}, MnemonicReader \citep{hu2018reinforced}, FusionNet \citep{huang2018fusionnet}, BERT \citep{Devlin2019BERTPO}, ALBERT \citep{lan2019albert}, RoBERTa \citep{Liu2019RoBERTaAR}, ELECTRA \citep{clark2020electra}, and SpanBERT \citep{joshi2020spanbert}.\footnote{See Appendix~\ref{appendix:modeling_approaches_implementation_details} for more details about the \modelingapproaches{} used to calculate concurrence.}

10 of our 20 modeling approaches are \emph{non-pretrained}. These approaches generally propose (1) better
sequence encoders for passages and questions (e.g., \citealp{lee2016learning,yang2017words,yu2018qanet}) and/or (2) improved attention mechanisms for question-passage interactions (e.g., \citealp{seo2017bidirectional,wang-etal-2017-gated,huang2018fusionnet}).

In contrast, the other 10 of our 20 modeling approaches are \emph{pre-trained}; these modeling approaches all use the Transformer architecture \citep{Vaswani2017AttentionIA}, but improve performance by proposing better pre-training procedures and objectives.
These pre-trained modeling approaches are generally evaluated on a suite of downstream tasks, in contrast to non-pretrained modeling approaches, which generally evaluate on a single benchmark.

All of these modeling approaches were originally evaluated on SQuAD, though several (e.g., SpanBERT) were also evaluated on other QA benchmarks.
We evaluate each \modelingapproach{} on each benchmark with the same training hyperparameters used for SQuAD, as well as \numHyperparameters{} additional randomly sampled hyperparameter settings.

\paragraph{Extractive QA benchmarks.} In this work, we study concurrence between three broad classes of extractive QA benchmarks: (i)~human-constructed, (ii)~cloze, and (iii)~synthetic. \Humanconstructedbenchmarks{} contain human-written natural language questions and passages; examples include SQuAD, NewsQA \citep{trischler-etal-2017-newsqa}, and NaturalQuestions \citep{Kwiatkowski2019NaturalQA}. On the other hand, cloze benchmarks (e.g., Children's Book Test or CNN; \citealp{hill2015goldilocks,hermann2015teaching}) contain cloze questions, which are ``fill-in-the-blank'' statements with masked answers. These questions are usually automatically-generated from human-written natural language passages. Finally, synthetic benchmarks contain programmatically-generated questions and passages (e.g., the bAbI task suite; \citealp{Weston2016TowardsAQ}).

\section{Do Modeling Improvements Hold Across Human-Constructed Benchmarks?}\label{sec:squad_vs_human_constructed}

\begin{table}
\resizebox{\columnwidth}{!}{%
\begin{tabular}{lccccc}
  \footnotesize
    & \rot[45][2em]{MRQA NewsQA}
    & \rot[45][2em]{MRQA NQ}
  & \rot[45][2em]{MRQA DROP}
  & \rot[45][2em]{MRQA HotpotQA}
  & \rot[45][2em]{QAMR} \\
    \midrule
    SQuAD  & 0.87 & 0.84  & 0.77 & 0.92 & 0.94 \\
    MRQA NewsQA & - & 0.82  & 0.83 & 0.92 & 0.87 \\
  MRQA NQ & 0.82 & -  & 0.69 & 0.80 & 0.80 \\
  MRQA DROP  & 0.83 & 0.69  & - & 0.79 & 0.83 \\
  MRQA HotpotQA & 0.92 & 0.80 & 0.79 & - & 0.89 \\
    \bottomrule
\end{tabular}
}
  \caption{Concurrence between human-constructed benchmarks. Despite differences in their crowdsourcing setup, passage and question distributions, and even linguistic phenomena of interest, human-constructed benchmarks generally have high concurrence ($\tau$) with each other on our testbed of modeling approaches.}
  \label{tab:human_constructed_heatmap}
\end{table}

\begin{table*}
  \centering
  \footnotesize
  \begin{tabular}{llllrrc@{}}
    \toprule
    \textbf{Benchmark} & \textbf{Question (Q)} & \textbf{Passage (P)} & \textbf{Phenomena of Interest} & \textbf{$|$Q$|$} & \textbf{$|$P$|$} &  \textbf{Q $\independent$ P} \\ \midrule
    SQuAD & Crowdsourced & Wikipedia & Predicate-Argument Structure & 11 & 137 & \xmark \\
    QAMR & Crowdsourced & Wikipedia & Predicate-Argument Structure & 7 &  25 & \xmark \\
    NewsQA & Crowdsourced & News articles & Predicate-Argument Structure & 8 & 599 & \cmark \\
    NaturalQuestions & Search logs & Wikipedia & Predicate-Argument Structure & 9 & 153 & \cmark \\
    HotpotQA & Crowdsourced & Wikipedia & Multi-Hop Reasoning & 22 & 232 & \xmark \\
    DROP & Crowdsourced & Wikipedia & Numerical Reasoning & 11 & 243  & \xmark \\
  \bottomrule
  \end{tabular}
  \caption{Differences between the various human-constructed benchmarks evaluated. $Q \independent P$ is true (\cmark) if the question was written independently from the associated passage. $|$Q$|$ and $|$P$|$ denote average question and passage token length, respectively.}
  \label{tab:human_constructed_comparison}
\end{table*}

\begin{figure*}
  \centering
  \includegraphics[width=\linewidth]{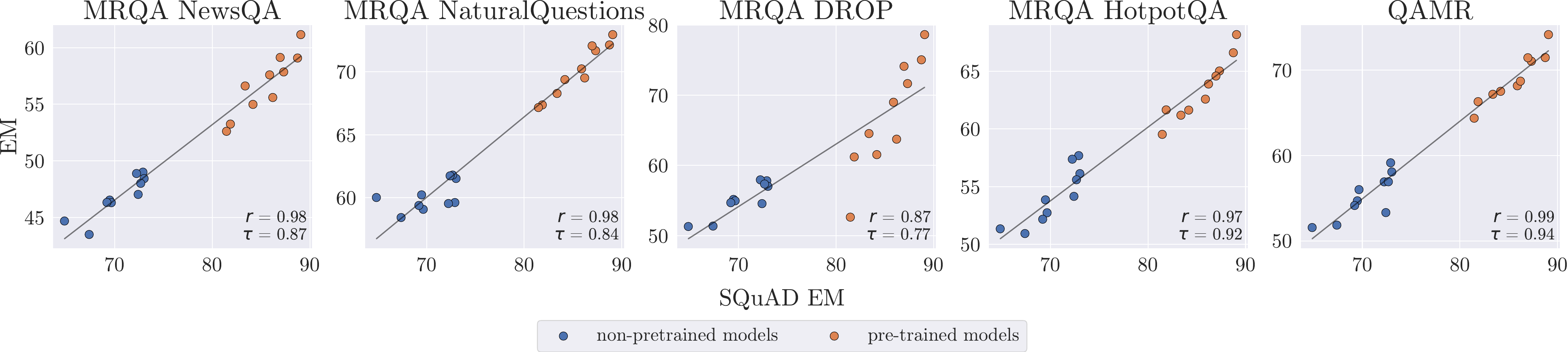}
  \caption{Human-constructed benchmarks have high concurrence with each other on both pre-trained and non-pretrained modeling approaches.}
  \label{fig:squad_vs_human_constructed}
\end{figure*}

Many extractive question answering benchmarks are \humanconstructed{}---they contain human-written natural language questions and passages. However, differences in the data collection procedure may yield benchmarks with dramatically different passage and question distributions. Do modeling improvements hold across benchmarks despite these differences?

\paragraph{Setup.}
We study the concurrence between six \humanconstructedbenchmarks{}: SQuAD, NewsQA, NaturalQuestions, DROP \citep{Dua2019DROP}, HotpotQA \citep{yang-etal-2018-hotpotqa}, and QAMR \citep{michael-etal-2018-crowdsourcing}.
We use the MRQA versions of NewsQA, NaturalQuestions, DROP, and HotpotQA \citep{fisch-etal-2019-mrqa}.
Table~\ref{tab:human_constructed_comparison} summarizes their high-level differences.
See Appendix~\ref{appendix:human_constructed_examples} for examples from  \humanconstructedbenchmarks{}.

\subsection{Results} \paragraph{\Humanconstructedbenchmarks{} have high concurrence amongst themselves.} Despite differences in benchmark crowdsourcing setups, passage and questions distributions, and even linguistic phenomena of interest, modeling improvements generally hold across human-constructed benchmarks (Table~\ref{tab:human_constructed_heatmap}). Furthermore, concurrence is high over both non-pretrained and pre-trained modeling approaches (Figure~\ref{fig:squad_vs_human_constructed}).

For example, SQuAD, NewsQA, and NaturalQuestions differ in their passage-question joint relationship. In SQuAD, crowdworkers are employed to write questions given Wikipedia passages, but this results in questions with high lexical overlap with salient passage sentences. To minimize such overlap in NewsQA, crowdworkers write questions given only bullet-point summaries of the passages, rather than the passages themselves. Finally, questions in NaturalQuestions are written independently of their provided passage.
These different crowdsourcing protocols drastically affect the ease and cost of benchmark construction, but SQuAD, NewsQA, and NaturalQuestions have high concurrence despite these differences.

\paragraph{Concurrence is high even when benchmarks focus on different phenomena.} We also see that MRQA DROP and MRQA HotpotQA have surprisingly high concurrence with other human-constructed benchmarks (e.g., SQuAD and NaturalQuestions), despite their relatively specialized focus on particular linguistic phenomena (numerical and multi-hop reasoning, respectively).\footnote{Note that MRQA DROP is a subset of the original benchmark that removes questions with non-extractive answers (e.g., answer is the result of an arithmetic operation).}
This suggests that modeling improvements on benchmarks that target general reasoning over predicate-argument structure also improve performance on benchmarks that focus on different phenomena.
We hypothesize this occurs because benchmarks are more similar than we'd otherwise expect (e.g., due to reasoning shortcuts; \citealp{min-etal-2019-compositional}), and better reasoning over predicate-argument structure may be generally useful for other phenomena of interest.

\section{Exploring the Limits of Concurrence}

Our results in \S\ref{sec:squad_vs_human_constructed} indicate that \humanconstructedbenchmarks{} have high concurrence with each other, despite differences in their phenomena of interest and passage and question distributions.
Just how different can a benchmark be, while maintaining high concurrence with human-constructed benchmarks?
In \S\ref{sec:squad_vs_subsampled_squad} we investigate the role of training dataset size on concurrence---while larger training datasets often yield better systems with higher end-task accuracy, are they necessary for comparing modeling approaches? 
In \S\ref{sec:squad_vs_cloze}, we measure concurrence between human-constructed and cloze benchmarks to better understand the role of human-written questions and passages in concurrence. Cloze benchmarks have high concurrence with human-constructed benchmarks, indicating that human-written questions and passages are not necessary for concurrence with human-constructed benchmarks.
To take this to an extreme, \S\ref{sec:babi} evaluates concurrence between programmatically-generated synthetic benchmarks (the bAbI task suite) with \humanconstructedbenchmarks{}. Our results show that the bAbI tasks have low concurrence with \humanconstructedbenchmarks{}.
Having found this breaking point, we work backwards to build a minimal benchmark with high concurrence, which will enable us to better understand sufficient conditions for concurrence.
In 
\S\ref{sec:squad_vs_synthetic_cloze}, we construct a benchmark that has no linguistic structure or complex reasoning but still has high concurrence with \humanconstructedbenchmarks{} over \nonpretrained{} models.
Finally, \S\ref{sec:squad_vs_wikidata} shows that a synthetic benchmark that requires richer reasoning between question and passage tokens can achieve high concurrence with \humanconstructedbenchmarks{} on \emph{both} pre-trained and \nonpretrained{} \modelingapproaches{}.

\subsection{Downsampling Benchmarks}
  \label{sec:squad_vs_subsampled_squad}

\begin{figure*}
  \centering
  \includegraphics[width=\textwidth]{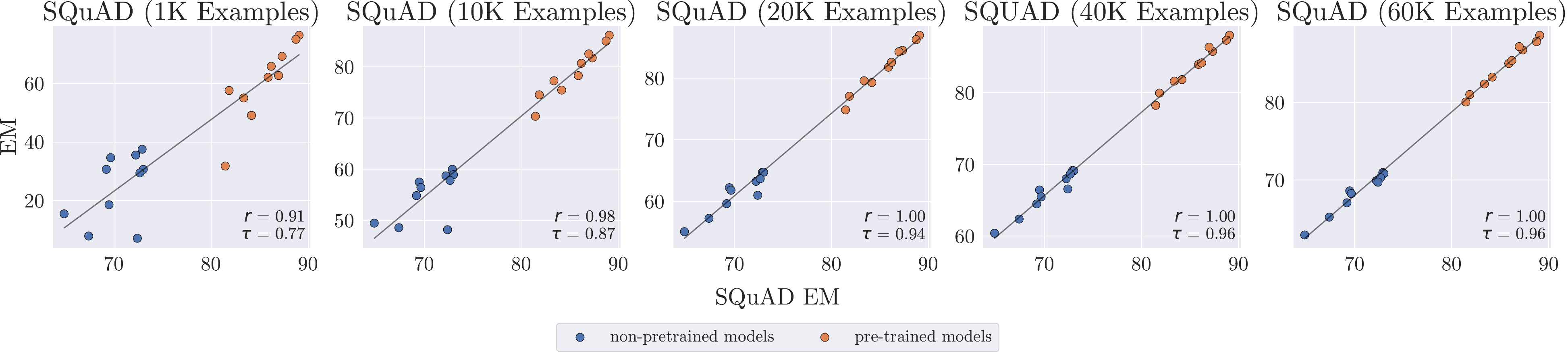}
  \caption{Downsampling the SQuAD training dataset can yield high concurrence with the full SQuAD benchmark on both pre-trained and non-pretrained modeling approaches. In particular, 10K training examples are sufficient for high concurrence on pre-trained models, and 20K examples yields high concurrence on non-pretrained mdoels.}
  \label{fig:benchmark_size}
\end{figure*}

Many existing \humanconstructed{} extractive QA benchmarks contain a large number of examples, increasing their cost of construction. For example, SQuAD has \numprint{87599} question-answer pairs in its training split. Are large training datasets necessary for comparing modeling approaches?

\paragraph{Setup.}
We study the extent to which subsamples of SQuAD concur with the full SQuAD benchmark (88K examples) and five other \humanconstructedbenchmarks{}.
We experiment with randomly generated subsets of the SQuAD training set with 1K, 10K, 20K, 40K, and 60K training examples. We use the original SQuAD development set ($\sim$10K examples) for evaluation.

\begin{table}
  \footnotesize
\resizebox{\columnwidth}{!}{%
  \begin{tabular}{lccccc}
    & \multicolumn{5}{c}{Downsampled SQuAD Size}\\
    \cmidrule(lr){2-6}
    & 60K
    & 40K
  & 20K
  & 20K
  & 1K \\
    \midrule
    SQuAD  & 0.96 & 0.96  & 0.94 & 0.87 & 0.77 \\
    MRQA NewsQA & 0.92 & 0.92  & 0.89 & 0.89 & 0.77 \\
  MRQA NQ & 0.84 & 0.84  & 0.81 & 0.78 & 0.63 \\
    \bottomrule
\end{tabular}
}
  \caption{
  Beyond a baseline threshold of 20K examples, downsampling the SQuAD training set minimally affects concurrence with the full SQuAD benchmark and other human-constructed benchmarks}
  \label{tab:benchmark_size_heatmap}
\end{table}

\paragraph{Results.} Downsampling the SQuAD training set from 88K to 20K examples does not substantially affect concurrence with the full SQuAD benchmark and other \humanconstructedbenchmarks{} (Table~\ref{tab:benchmark_size_heatmap}).
Concurrence is high on both non-pretrained and pre-trained modeling approaches (Figure~\ref{fig:benchmark_size}).
Downsampling to 10K examples slightly reduces concurrence with non-pretrained modeling approaches. Concurrence with pre-trained models only begins to degrades when using 1K training examples, indicating that few-shot settings are likely categorically different and worth studying separately.

\subsection{Cloze Benchmarks}\label{sec:squad_vs_cloze}

\begin{figure*}
  \centering
  \includegraphics[width=\textwidth]{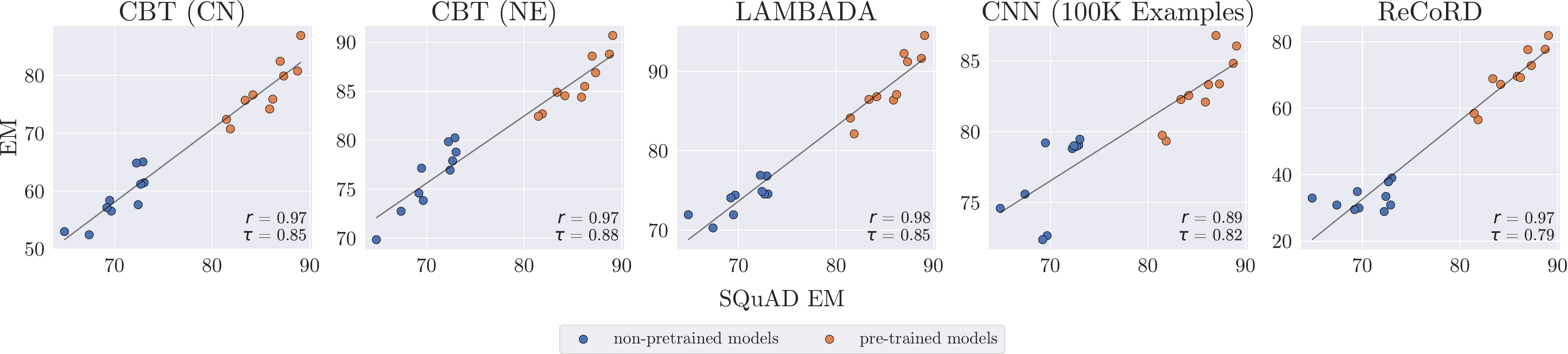}
  \caption{Despite their cloze-formatted question, which differ from questions typically found in \humanconstructedbenchmarks{}, cloze benchmarks can have high concurrence with SQuAD (CBT-CN, CBT-NE, and LAMBADA), though this is not always the case (CNN, ReCoRD).}
  \label{fig:squad_vs_cloze}
\end{figure*}

To better understand the importance of human-written questions and passages, we measure concurrence between \humanconstructedbenchmarks{} and cloze benchmarks. Cloze extractive question answering benchmarks contain cloze questions, which are ``fill-in-the-blank'' statements with masked answers. Large cloze benchmarks are cheap to construct because examples can be automatically generated by eliding spans from naturally-occurring text. Although the passages in cloze benchmarks are natural language, their fill-in-the-blank require more guessing from context, rather than the answer deduction typically found in human-constructed benchmarks.

\paragraph{Setup.} We study the Children's Book Test (CBT; \citealp{hill2015goldilocks}), LAMBADA \citep{paperno2016lambada}, CNN \citep{hermann2015teaching}, and ReCoRD \citep{zhang2018record} cloze benchmarks and measure their concurrence with \humanconstructedbenchmarks{} on our testbed of modeling approaches.
We follow prior work \citep{dhingra2017gated} and evaluate on subsets of CBT where the answer token is either a common noun (CBT-CN) or a named entity (CBT-NE). In addition, we use a subsampled version of the CNN benchmark with 100K training examples to save compute.
See Appendix~\ref{appendix:cloze_examples} for examples from the cloze benchmarks we study.

\paragraph{Results.} Despite using programmatically-generated cloze questions, cloze benchmarks (e.g., CBT and LAMBADA) can have high concurrence with \humanconstructedbenchmarks{} (Table~\ref{tab:cloze_heatmap}).
On the other hand, CNN and ReCoRD have lower concurrence with \humanconstructedbenchmarks{}, especially on non-pretrained modeling approaches---the performance improvements between pre-trained modeling approaches are still largely preserved (Figure~\ref{fig:squad_vs_cloze}).

\begin{table}
\resizebox{\columnwidth}{!}{%
\begin{tabular}{lccccc}
  \footnotesize
    & \rot[45][2em]{CBT (CN)}
    & \rot[45][2em]{CBT (NE)}
  & \rot[45][2em]{LAMBADA}
  & \rot[45][2em]{CNN (100K)}
  & \rot[45][2em]{ReCoRD} \\
    \midrule
    SQuAD  & 0.85 & 0.88  & 0.85 & 0.82 & 0.79 \\
    MRQA NewsQA & 0.92 & 0.93  & 0.87 & 0.76 & 0.77 \\
  MRQA NQ & 0.78 & 0.79 & 0.75 & 0.79 & 0.88 \\
    \bottomrule
\end{tabular}
}
  \caption{Concurrence between programmatically-generated cloze benchmarks and human-constructed benchmarks can be high (e.g., CBT and LAMBADA), but not always (CNN and ReCoRD).}
  \label{tab:cloze_heatmap}
\end{table}

Concurrence on CNN is lower due to a pair of outlier modeling approaches---DocumentReader, with and without external linguistic features. We hypothesize that these models do poorly on CNN because some aspects of their preprocessing are SQuAD-specific; this may have also influenced architecture design.
ReCoRD's low overall concurrence comes from the poor performance of non-pretrained \modelingapproaches{}. This may be due to ReCoRD's construction procedure, since a filtering step removed all examples that were correctly answered by a strong \nonpretrained{} \modelingapproach{} (SAN, with SQuAD dev. EM of 76.24; \citealp{liu2018stochastic}). ReCoRD has low concurrence with SQuAD on \modelingapproaches{} that are weaker than SAN, and high concurrence on \modelingapproaches{} that outperform SAN.

\subsection{High Concurrence Is Not Universal: Improvements Do Not Hold On bAbI}\label{sec:babi}

\begin{figure*}
  \centering
  \includegraphics[width=\textwidth]{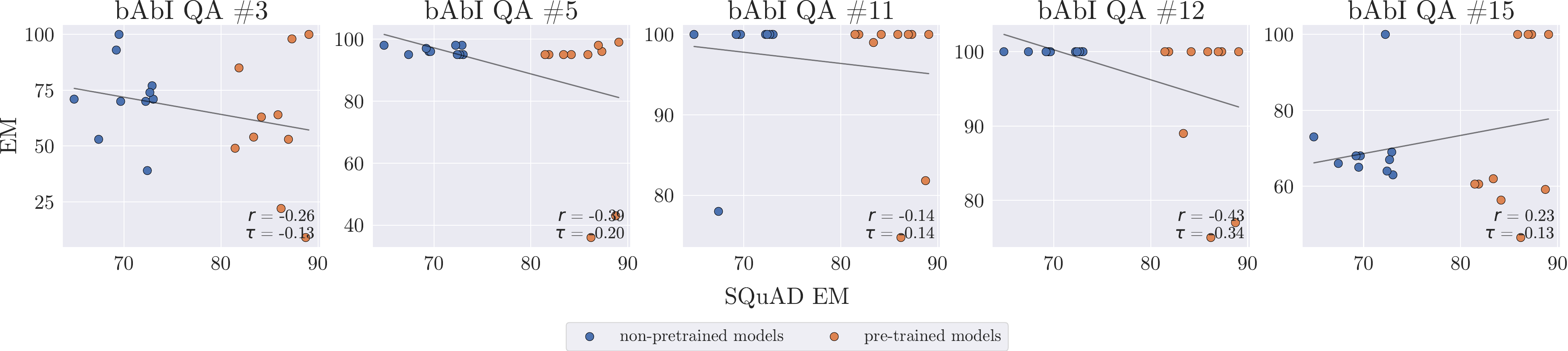}
  \caption{Modeling approaches perform perfectly or at near-chance performance on the bAbI tasks, limiting their ability to recapitulate historical findings on SQuAD (see Appendix~\ref{appendix:existing_synthetic_benchmarks_full_results} for the full results on all tasks).}
  \label{fig:squad_vs_synthetic}
\end{figure*}

Having established that human-written passages are not necessary for high concurrence with \humanconstructedbenchmarks{} (\S\ref{sec:squad_vs_cloze}), we take this to an extreme by evaluating concurrence between \humanconstructedbenchmarks{} and synthetic extractive question answering benchmarks, which contain questions and passages that are programmatically generated (and possibly not even natural language).
The bAbI task suite contains 20 synthetic question-answering benchmarks, each of which focuses on a particular skill required by a competent dialogue system (e.g., fact retrieval, subject-object relations, counting).
The textual data is generated from a simulated toy environment.

\paragraph{Setup.} We consider the 11 tasks that can be losslessly converted to an extractive format (Tasks 1, 2, 3, 4, 5, 11, 12, 13, 14, 15, 16). For each task, we use the two officially-released data settings: one setting has \numprint{900} training examples and 100 development examples, and the other has \numprint{9000} training examples and \numprint{1000} development examples. In this section, we focus on the setting with \numprint{900} training examples, since all modeling approaches do nearly perfectly on almost all tasks with \numprint{9000} examples (Appendix~\ref{appendix:existing_synthetic_benchmarks_full_results}).
See Appendix~\ref{appendix:synthetic_examples} for examples from the existing synthetic benchmarks we study.

\paragraph{Results and Discussion.} The bAbI tasks have low concurrence with \humanconstructedbenchmarks{}---high concurrence is not universal. Modeling approaches often have either near-perfect or near-random performance (Figure~\ref{fig:squad_vs_synthetic}).

\subsection{What is Sufficient for Concurrence on Non-Pretrained Modeling Approaches?}\label{sec:squad_vs_synthetic_cloze}

To better understand the sufficient conditions for concurrence with \humanconstructedbenchmarks{}, we are interested in constructing a minimal synthetic benchmark with high concurrence.
Given that human-written passages and questions are not necessary for high concurrence with \humanconstructedbenchmarks{} (\S\ref{sec:squad_vs_cloze}), but the programmatically-generated bAbI synthetic benchmarks have low concurrence (\S\ref{sec:babi}), we design a minimal synthetic benchmark with high concurrence with \humanconstructedbenchmarks{} over non-pretrained modeling approaches.

\paragraph{Setup.} Questions in extractive QA benchmarks can often be answered by exploiting lexical overlap between question and passage tokens \citep{weissenborn2017making,Krishna2019ThievesOS}. To better understand the limits of concurrence, we build a minimal synthetic cloze benchmark (\fuzzypmsyntheticqa{}) that explicitly targets this fuzzy pattern-matching and find that it has high concurrence with SQuAD on non-pretrained modeling approaches.
Figure~\ref{fig:synthetic_patternmatching_example} shows a sample passage and question-answering pairs. We use \numprint{10000} questions for training and \numprint{10000} questions for evaluation.
See Appendix~\ref{appendix:synthetic_patternmatching_construction_details} for further details about \fuzzypmsyntheticqa{}'s construction.

\begin{figure}
\begin{framed}
\footnotesize
  \textbf{Passage Snippet:}
  \emph{
    ...
    chests Melchior divorced might whereof 37th Kadima milling raved Salib melanocephala Pilgrims \hlc[question1]{chop} Prosser draftsmanship 203 Caesarius madam Deconstruction Guevara Amalia
    ...
  }

  \textbf{Question:} \emph{\hlc[question1]{Pigs corncrake XXXXX 286 airmanship Kition gracious Modernism Raul}}

  \textbf{Answer:} \emph{chop}

\end{framed}
\caption{Example passage and question-answer pair from \fuzzypmsyntheticqa.
}
\label{fig:synthetic_patternmatching_example}
\end{figure}

\paragraph{Passage Generation.} We generate the passage by randomly sampling 150 tokens from the uniform distribution over a token vocabulary. The token vocabulary is taken from the WikiText-2 training set \citep{merity2017pointer} and has \numprint{68429} types.

\paragraph{Answer Generation.} The answer token is randomly selected from the generated passage.

\paragraph{Cloze Question Generation.} To generate the cloze question, we first extract the answer token's local context (up to 10 tokens) and mask out the answer token. Then, we corrupt the cloze question by (1)~randomly replacing its tokens with related tokens (100 approximate nearest neighbor tokens in the vocabulary, measured by vector distance in the \pretrained{} English FastText embeddings), (2)~locally permuting its tokens (within 3 positions), and (3)~applying word dropout (with rate 0.2).

\begin{figure}
  \centering \includegraphics[width=\columnwidth]{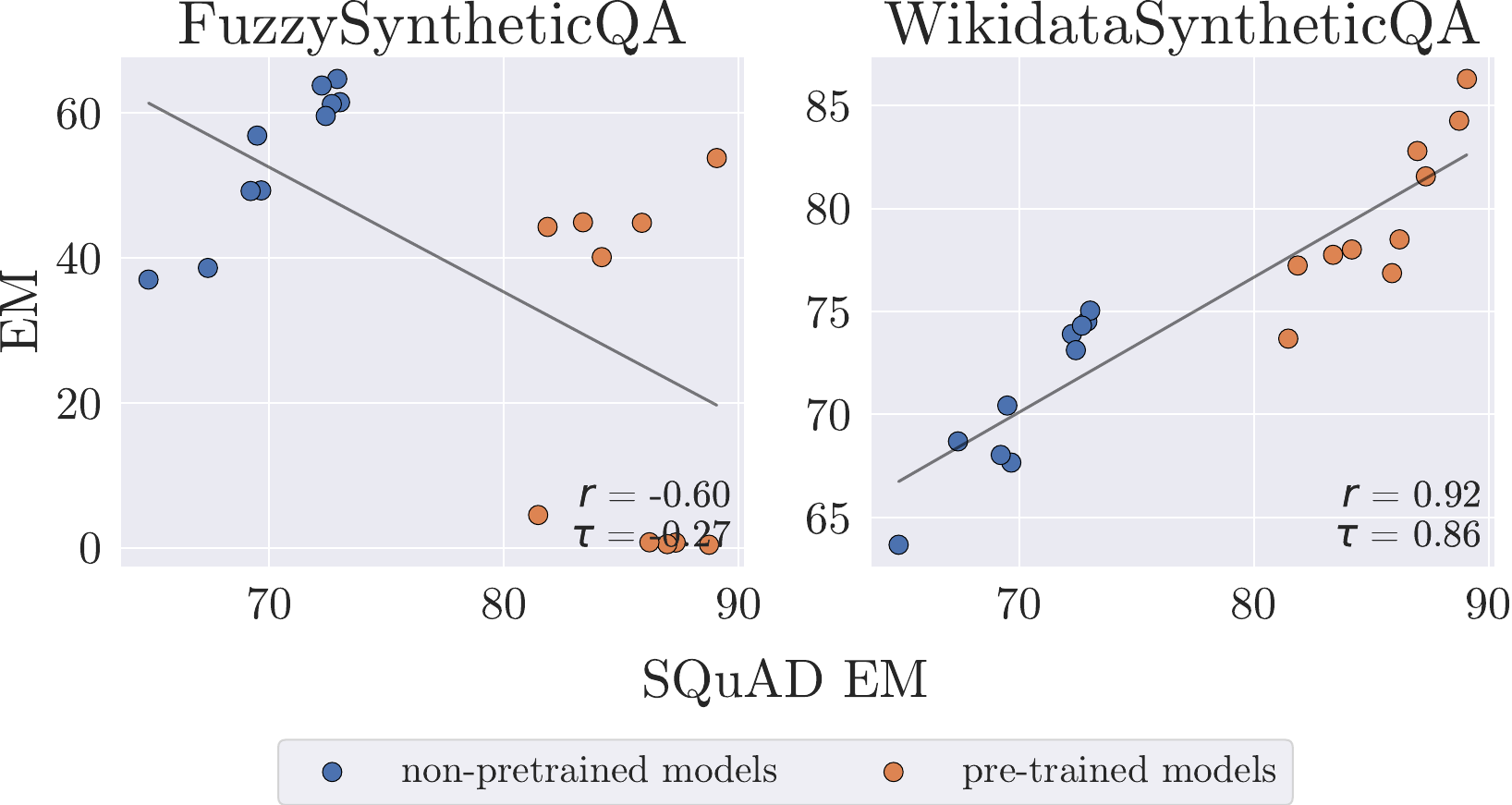}
  \caption{\textbf{Left:} \fuzzypmsyntheticqa has high concurrence with SQuAD on \nonpretrained{} \modelingapproaches{}, but \pretraining{} does not increase performance, leading to low overall concurrence. \textbf{Right:} Despite lacking natural language structure, \wikidatasyntheticqa has high concurrence with SQuAD.}
  \label{fig:squad_vs_synthetic_combined}
\end{figure}

\paragraph{Results and Discussion.} \fuzzypmsyntheticqa{} has high concurrence with \humanconstructedbenchmarks{}, but only on \nonpretrained{} \modelingapproaches{}---concurrence on pre-trained \modelingapproaches{} is much lower (Figure~\ref{fig:squad_vs_synthetic_combined}). Even benchmarks that lack much linguistic structure can have high concurrence with \humanconstructedbenchmarks{}, as long as they require similar phenomena (in this case, fuzzy lexical matching between the question and passage).

Why do improvements in pre-training not hold on \fuzzypmsyntheticqa? One potential reason is that passages in \fuzzypmsyntheticqa{} lack of linguistic structure.
To evaluate this hypothesis, we generate \fuzzypmsyntheticqa{} questions from English Wikipedia passages, rather than sampling from the uniform distribution over tokens, but this still results in low concurrence with human-constructed benchmarks on \pretrained{} \modelingapproaches{} ($r = -0.49$, $\tau = -0.19$), indicating that the low concurrence comes from more than just a lack of natural language passages (Appendix~\ref{appendix:synthetic_fuzzy_pattern_matching_benchmarks_full_results}).

\subsection{What is Sufficient for Concurrence on Pre-Trained and Non-Pretrained Modeling Approaches?}\label{sec:squad_vs_wikidata}

Having found a minimal synthetic benchmark that achieves high concurrence with \humanconstructedbenchmarks{} on non-pretrained modeling approaches (\S\ref{sec:squad_vs_synthetic_cloze}), we show that a synthetic benchmark that requires richer reasoning between question and passage tokens is sufficient for high concurrence on \emph{both} non-pretrained and pre-trained modeling approaches.

\paragraph{Setup.} We construct \wikidatasyntheticqa{}, a benchmark derived from \wikidata{} triples; Figure~\ref{fig:synthetic_wikidata_example} shows a sample passage and question-answering pairs.
Knowledge graphs like \wikidata{} are rich sources of complex relations between entities, which enables us to increase the complexity of question-passage token relations beyond the simple noising and corruptions of \fuzzypmsyntheticqa{}.
We use \numprint{10000} questions for training and \numprint{9835} question-answer pairs for evaluation.
See Appendix~\ref{appendix:wikidata_construction_details} for further details about \wikidatasyntheticqa{}'s construction.

\begin{figure}
\begin{framed}
\footnotesize
\textbf{Passage Snippet:}
\emph{
  Mae Jemison profession astronaut . STS-47 orbits completed \hlc[question1]{126.0} . STS-47 crew member Mae Carol Jemison . Mae Jemison worked for NASA . \hlc[question2]{Mae C. Jemison} award received Rachel Carson Award . Mae Jemison \hlc[question3]{birthplace} Decatur .
  ...
}

  \textbf{Question:} \emph{\hlc[question1]{human spaceflight orbits completed XXXXX}}

  \textbf{Answer:} \emph{126.0}

  \textbf{Question:} \emph{\hlc[question2]{Rachel Carson Award honor received by XXXXX}}

  \textbf{Answer:} \emph{Mae C. Jemison}

  \textbf{Question:} \emph{\hlc[question3]{Human XXXXX The River City}}

  \textbf{Answer:} \emph{birthplace}
\end{framed}
\caption{Example passage and question-answer pairs from \wikidatasyntheticqa.
}
\label{fig:synthetic_wikidata_example}
\end{figure}

\paragraph{Wikidata Background.} \wikidata{} is a knowledge graph connecting entities via relations. \wikidata{} entities and relations include a \emph{label}, the most common name that an entity is known by, and \emph{aliases}, alternative names for entities.
For example, the entity \texttt{Mae\_C.\_Jemison} has the label \textit{``Mae C. Jemison''}, with aliases \textit{``Mae Jemison''} and \textit{``Mae Carol Jemison''}. 
We treat labels and aliases as potential surface realizations of entities and relations.

\paragraph{Generation Preliminaries.} Generating a passage requires a set of \wikidata{} triples. To select these triples, we first randomly choose a seed entity from the \numprint{10000} \wikidata{} entities with the highest PageRank score \citep{ilprints422}.
We then extract the triples from the seed entity and all entities connected to the seed entity. Finally, we randomly sample 50 triples for use in generation.

\paragraph{Passage Generation.}
Given the set of 50 \wikidata{} triples, we realize triples into textual surface forms by selecting a random \wikidata{} label or alias for each triple element.
The final passage is formed by concatenating the realizations of all triples and adding a delimiter token between them to mimic sentential structure.

\paragraph{Answer Generation.} We generate an answer span by selecting a random triple used in the passage generation process, and then choosing a random element of that triple. The passage realization of this random element is the answer span.

\paragraph{Cloze Question Generation.} To generate the cloze question, we take the triple used for answer generation and mask out the particular element marked as the answer. We realize the non-answer triple elements into textual forms by selecting a random \wikidata label or alias for each triple element.
Then, we optionally and randomly replace the predicate with its inverse (if one exists), reversing the subject and the object to maintain consistency.
We also optionally and randomly replace the remaining unmasked entity (i.e., the triple subject or object that was not masked) with one of its hypernyms, challenging models' knowledge of such relations.

\paragraph{Results and Discussion.} As Figure~\ref{fig:squad_vs_synthetic_combined} shows, \wikidatasyntheticqa{} has high concurrence with \humanconstructedbenchmarks{}, despite its lack of natural language passages or questions.

We hypothesize that \wikidatasyntheticqa{} has higher concurrence with \humanconstructedbenchmarks{} than \fuzzypmsyntheticqa{} because correctly answering its examples often requires reasoning about hypernymy relations between entities and inverse relations between predicates---it is conceivable that \pretrained{} modeling approaches are better-equipped to handle and use these lexical relations.
In addition, the \wikidata{} aliases provide sufficient lexical variation such that the benchmark is not trivially solvable through string pattern-matching (removing aliases from the generation procedure results in near-perfect performance from all modeling approaches). In contrast, high performance on \fuzzypmsyntheticqa{} simply requires matching similar tokens in the passage and question---models can achieve high performance by simply learning the similarity relationships in the FastText vector space.

\section{Related Work}

A recent line of work examines whether \emph{systems} have overfit to particular test sets by taking existing systems and evaluating them on newly-constructed test sets \citep{recht2019imagenet,yadav2019cold,miller2020effect}. Recent work has also studied whether higher-performing systems are more robust by studying the correlation between in-domain and out-of-domain improvements \citep{taori2020measuring,djolonga2020robustness}.

In contrast, this work examines whether improvements from \emph{\modelingapproaches{}} hold across benchmarks. We train and test \modelingapproaches{} on a variety of existing and newly-constructed benchmarks. In this regard, our work is similar to the study of \citet{kornblith2019better}, who find that performance improvements on ImageNet are well-correlated with performance improvements on other benchmarks.

\section{Conclusion}

This work studies whether QA modeling improvements hold across the diverse landscape of QA benchmarks.
We develop the notion of \emph{concurrence}, which quantifies the similarity between benchmarks' rankings of modeling approaches.
Experiments with \numBenchmarks{} QA benchmarks and \numModels{} diverse \modelingapproaches{} indicate that human-constructed benchmarks largely have high concurrence amongst themselves, even when their passage and question distributions or linguistic phenomena of focus are very different.
To better understand how different benchmark attributes affect concurrence, we explore downsampled benchmarks and various programmatically-generated benchmarks, the latter having high concurrence only when they target phenomena that are also useful for better performance on \humanconstructedbenchmarks{} (e.g., identifying paraphrase and lexical overlap).
Our results indicate that the modeling improvements studied hold broadly, despite years of intense community focus on a small number of benchmarks.

\section*{Acknowledgements}

We thank the anonymous reviewers for their feedback and comments that helped improve this work.
NL was supported by an NSF Graduate Research Fellowship under grant number DGE-1656518. Other funding was provided by a PECASE Award.

\section*{Limitations}

While we conducted an extensive set of experiments to gain a broad picture of whether modeling improvements hold between benchmarks, it is always possible to investigate more settings. While our study covers a representative set of \numModels{} non-pretrained and pre-trained modeling approaches, it is conceivable that evaluating more modeling approaches (or a different set of modeling approaches) on additional benchmarks (or a different set of benchmarks) would have led to different results.

Furthermore, although we evaluate each \modelingapproach{} on each benchmark with the same training hyperparameters used for SQuAD, as well as \numHyperparameters{} additional randomly sampled hyperparameter settings (\numModels{}~\texttimes~\numBenchmarks{}~\texttimes~6 = 3840 experiments in total), it is possible that the SQuAD hyperparameters for some \modelingapproaches{} happen to be more general than other \modelingapproaches{}. Ideally, each \modelingapproach{} would be individually tuned to maximize performance on every benchmark, but doing so requires prohibitive amounts of compute and researcher effort---we believe that our experiments have enough coverage with respect to hyperparameter optimization.

\bibliography{custom}
\bibliographystyle{acl_natbib}
\clearpage
\onecolumn
\begin{appendices}
\section{Implementation Details of \ModelingApproaches{} Evaluated}\label{appendix:modeling_approaches_implementation_details}

We evaluated a representative subset of \numModels{} extractive question answering \modelingapproaches{}, published between 2016 to 2020 (Table~\ref{tab:models}). Below, we describe implementation details for all the \modelingapproaches{} evaluated.

\begin{table}[ht]
\centering
\footnotesize
\begin{tabular}{m{3.1cm}>{\centering\arraybackslash}rr}
\toprule
\multirow{2}{*}{Modeling Approach} & \multicolumn{2}{c}{SQuAD 1.1 Dev. EM}\\
\cmidrule{2-3}
{} &  Our Reproduction & Published \\
\midrule
RaSoR                                     &              64.9 &      66.4 \\
BiDAF                                     &              67.4 &      67.7 \\
DocumentReader                            &              69.7 &      69.5 \\
DocumentReader \newline(no external features)     &              69.2 &       - \\
BiDAF++                                   &              69.5 &      71.6 \\
MnemonicReader                            &              73.0 &      71.8 \\
MnemonicReader \newline(no external features)     &              72.7 &       - \\
QANet                                     &              72.4 &      73.6 \\
FusionNet                                 &              72.9 &      75.0 \\
FusionNet \newline(no external features)          &              72.2 &       - \\
\midrule
BERT (base, uncased)                      &              81.5 &      80.8 \\
BERT (large, uncased)                     &              84.2 &      84.1 \\
BERT (large, uncased, whole-word masking) &              87.3 &      86.7 \\
ALBERT (base, V1)                         &              81.9 &      82.3 \\
ALBERT (xxlarge, V1)                      &              89.1 &      89.3 \\
RoBERTa (base)                            &              83.4 &         - \\
RoBERTa (large)                           &              87.0 &      88.9 \\
ELECTRA (base)                            &              85.9 &      84.5 \\
SpanBERT (base)                           &              86.2 &         - \\
SpanBERT (large)                          &              88.7 &      88.1 \\
\bottomrule
\end{tabular}
\caption{Published and reproduced SQuAD 1.1 EM of all \numModels{} modeling approaches used for assessing concurrence. ``-'' indicates that the modeling approach has no published SQuAD 1.1 EM result.}
\label{tab:models}
\end{table}

\paragraph{RaSoR}\label{appendix:rasor_details}
We reimplement the RaSoR model of \citep{lee2016learning} with PyTorch in the AllenNLP \citep{allennlp} framework, following the original paper as closely as possible. While the authors released an implementation of their method (\texttt{\href{https://github.com/shimisalant/rasor}{github.com/shimisalant/rasor}}), the codebase is in Theano and inexplicably fails on passages that are significantly longer than those found in SQuAD (e.g., those found in the CNN benchmark).

\paragraph{BiDAF}\label{appendix:bidaf_details}
We use the reimplementation of BiDAF \citep{seo2017bidirectional} found in AllenNLP \citep{allennlp}.

\paragraph{DocumentReader (with and without external features)}\label{appendix:documentreader_details}
We use an reimplementation of DocumentReader \citep{chen2017reading} released at \texttt{\href{https://github.com/felixgwu/FastFusionNet}{github.com/felixgwu/FastFusionNet}}. The original DocumentReader approach uses external features from a part-of-speech tagger and named entity recognition system. To fairly compare to systems that do not use such external resources, we also run the models without these features. We keep the hand-crafted term-frequency and token exact match features defined in the DocumentReader paper.

We also make some changes to the DocumentReader preprocessing code.  In particular, the original implementation (\texttt{\href{https://github.com/facebookresearch/DrQA}{github.com/facebookresearch/DrQA}}) of these two modeling approaches (intended for training and evaluation on SQuAD) replaces all tokens without a \pretrained{} GloVe embedding (trained on 840B tokens from the Common Crawl) with a special unknown token---the reimplementation we use adopts the same practice. This preprocessing assumption works well for SQuAD, since the vast majority of SQuAD tokens also appear in the GloVe vocabulary. However, this preprocessing assumption does not apply to CNN---many of the special \texttt{@entity\emph{N}} and \texttt{@placeholder} markers, which anonymize entities to prevent models from deriving answers from world knowledge, are not in the GloVe vocabulary. As a result, the original DocumentReader implementation maps them all to a single unknown token, effectively preventing the model from telling valid answer choices apart and yielding a model that performs no better than the majority baseline. Keeping these special tokens in the model's vocabulary enables differentiating between different entities in a passage, which naturally improves performance (and are the reported numbers)---however, the modeling approaches' improvements on SQuAD still do not transfer to CNN.

\paragraph{BiDAF++}\label{appendix:bidafplusplus_details}
We modify an AllenNLP \citep{allennlp} reimplementation of the BiDAF++ \citet{clark-gardner-2018-simple} model originally used in pair2vec \citep{joshi-etal-2019-pair2vec} for evaluation on SQuAD 2.0 \citep{rajpurkar2018know}.

\paragraph{MnemonicReader}\label{appendix:mnemonicreader_details}
We use an reimplementation of MnemonicReader (\citealp{hu2018reinforced}; note the specific arXiv revision) released at \texttt{\href{https://github.com/HKUST-KnowComp/MnemonicReader}{github.com/HKUST-KnowComp/MnemonicReader}}. In particular, the reimplementation is of the vanilla MnemonicReader without reinforcement learning.

\paragraph{QANet}\label{appendix:qanet_details}
We use the reimplementation of QANet \citep{yu2018qanet} found in AllenNLP \citep{allennlp}. This reimplementation was used as a baseline method for DROP \citep{Dua2019DROP}.

\paragraph{FusionNet}\label{appendix:fusionnet_details}
We use an reimplementation of FusionNet \citep{chen2017reading} released at \texttt{\href{https://github.com/felixgwu/FastFusionNet}{github.com/felixgwu/FastFusionNet}}. This reimplementation was used as a baseline in \citet{wu2019fastfusionnet}. Drawing inspiration from DocumentReader, the FusionNet approach also uses external features from a part-of-speech tagger and named entity recognition system. As a result, we also run the models without these features to fairly compare to systems that do not use such external resources. We keep the hand-crafted term-frequency and token exact match features originally used in the FusionNet paper.

\paragraph{BERT (base, large, and wwm)}\label{appendix:bert_details}
We use the HuggingFace Transformers \citep{wolf-etal-2020-transformers} library to fine-tune BERT \citep{Devlin2019BERTPO} on extractive question answering benchmarks. In particular, we use the base, uncased, BERT \pretrained{} model, the large, uncased, BERT \pretrained{} model, and the large, uncased, BERT model \pretrained{} with whole-word masking.

\paragraph{ALBERT (base and xxlarge)}\label{appendix:albert_details}
We use the HuggingFace Transformers \citep{wolf-etal-2020-transformers} library to fine-tune ALBERT \citep{lan2019albert} on extractive question answering benchmarks. In particular, we use the base and xxlarge V1 ALBERT \pretrained{} models.

\paragraph{RoBERTa (base and large)}\label{appendix:roberta_details}
We use the HuggingFace Transformers \citep{wolf-etal-2020-transformers} library to fine-tune RoBERTa \citep{Liu2019RoBERTaAR} on extractive question answering benchmarks. In particular, we use the base and large RoBERTa \pretrained{} models.

\paragraph{ELECTRA (base)}\label{appendix:electra_details}
We use the HuggingFace Transformers \citep{wolf-etal-2020-transformers} library to fine-tune the ELECTRA base discriminator \citep{clark2020electra} on extractive question answering benchmarks.

\paragraph{SpanBERT (base and large)}\label{appendix:spanbert_details}
We use the author-released codebase (\texttt{\href{https://github.com/facebookresearch/SpanBERT}{github.com/facebookresearch/SpanBERT}}) to fine-tune SpanBERT \citep{joshi2020spanbert} on extractive question answering benchmarks. In particular, we use the base and large SpanBERT \pretrained{} models.

\section{Preprocessing Existing Benchmarks}\label{appendix:benchmark_preprocessing}

\subsection{Existing Human-Constructed Benchmarks}

We use the MRQA NewsQA, MRQA DROP, and MRQA HotpotQA benchmarks exactly as released by the MRQA 2019 shared task \citep{fisch-etal-2019-mrqa}.
The passages in MRQA NaturalQuestions contain HTML entities (e.g., \texttt{<P>} and \texttt{</P>}). The tokenizers used in \nonpretrained{} models frequently split these entities into separate tokens. For example, \texttt{<P>} may become \texttt{<}, \texttt{P}, and \texttt{>}. This is problematic because the entities are quite common in passages, and expanding them during tokenization drastically increases the passage lengths, which some \nonpretrained{} modeling approaches cannot handle due to GPU memory limits.
HTML entities are tokenized like this because they contain non-alphanumeric characters. As a result, we normalize HTML entities by replacing the non-alphanumeric characters. For example, \texttt{<P>} becomes \texttt{BPB}, and \texttt{</P>} becomes \texttt{EEPE}. These tokens are correctly kept intact. It's possible that modeling approaches that use subword information will perform worse with these normalized HTML entities, but we empirically observe that this normalization does not have a measurable impact on model performance.

QAMR questions were originally collected at the sentence level, but we concatenate these sentences to reconstruct the original passages they were sourced from. We then pair these reconstructed passages with the original QAMR questions.
It's possible for questions to become unanswerable at the passage-level. One case of his happens when two sentences have the same question---we filter out such questions that are asked for multiple sentences in a reconstructed passage.
Questions can also become unanswerable if relations between entities change between sentences. For example, given the passage \texttt{``Bill lived in California in 1920. Bill lived in Washington in 1921.''}, the question ``Where did Bill live'' is answerable within the context of a particular sentence, but not in the context of the entire passage. Manual examination of generated QAMR passages and questions suggests that this case is rather uncommon, but it may still introduce a small amount of noise into the benchmark.

\subsection{Existing Cloze Benchmarks}

To convert the CBT and CNN benchmarks to extractive format, we take the passages and question as-is. The answer span is designated as the first occurrence of the answer token in the passage.
To convert LAMBADA into extractive format, we follow the setup of \citet{cheng2020attending}.
The ReCoRD benchmark is used as-is, since it includes span-level annotations of answer tokens in passages.

\subsection{Existing Synthetic Benchmarks}

We consider tasks 1, 2, 3, 4, 5, 11, 12, 13, 14, 15, 16. The other tasks cannot be converted to extractive format (e.g., they require ``yes''/``no'' answers that do not appear in passages).
To convert the tasks in the bAbI benchmark to extractive format, we take the passages and question as-is. While the bAbI benchmark does not provide character-level span annotations for answers, questions come with ``supporting facts''---sentences in the passage that contain the answer. Thus, choose the first occurrence of the answer token in the supporting fact sentence as our answer span.

Some of the bAbI tasks, while usable in an extractive format in theory, cannot be trivially converted to the extractive format via the procedure above because the released benchmark's annotations do not appear in the passage. For instance, consider Figure~\ref{fig:babi_plural_example}, which shows an example drawn from the training set of Task 15. The answer provided in the benchmark is \texttt{``cat''}, although this token never appears in the passage---instead, \texttt{``cats''} does. In cases where the originally-labeled answer cannot be found in the supporting fact, but its pluralization is present, we use the pluralized answer as our answer span.

\begin{figure}[!ht]
\begin{framed}
\footnotesize
  \textbf{Passage:}
  \emph{Mice are afraid of cats. Gertrude is a mouse. Emily is a mouse. Wolves are afraid of sheep. Winona is a wolf. Jessica is a mouse. Cats are afraid of sheep. Sheep are afraid of cats.}\\

  \textbf{Question:} \emph{What is jessica afraid of?}

  \textbf{Answer:} \emph{cat}
\end{framed}
\caption{}\label{fig:babi_plural_example}
\end{figure}

\clearpage
\newpage

\section{Examples From Existing Benchmarks}\label{appendix:existing_examples}

\subsection{Examples From Existing \HumanConstructedBenchmarks{}}\label{appendix:human_constructed_examples}

Table~\ref{tab:existing_human_constructed_benchmark_examples} shows examples from the existing \humanconstructedbenchmarks{} we study.

\begin{table}[!h]
\centering
\footnotesize
\begin{tabular}{L{2.75cm}p{7.25cm}p{2.8cm}L{1.25cm}}
\toprule
{\bf Benchmark} & {\bf Passage} (some parts shortened with \texttt{...}) & {\bf Question} & {\bf Answer} \\
 \midrule
  MRQA NewsQA & \texttt{(CNET)  -- When Facebook Chief Executive Mark Zuckerberg recently announced a ``Like'' button that publishers could place on their Web pages, he predicted it would make the Web smarter and ``more social''. What Zuckerberg didn't point out is that widespread use of the Like button allows Facebook to track people as they switch from CNN.com to Yelp.com to ESPN.com, all of which are sites that have said they will implement the feature...} & \texttt{What does the like button allow?} & \texttt{Facebook to track people} \\
  \midrule
  MRQA NaturalQuestions & \texttt{BPB A shooting schedule is a project plan of each day 's shooting for a film production . It is normally created and managed by the assistant director , who reports to the production manager managing the production schedule . Both schedules represent a timeline stating where and when production resources are used . EEPE} & \texttt{who 's job is it to schedule each day 's shooting} & \texttt{assistant director} \\
  \midrule
  MRQA DROP & \texttt{Coming off their win over the Chargers, the Bills flew to Dolphin Stadium for a Week 8 AFC East duel with the Miami Dolphins.  In the first quarter, Buffalo trailed early as Dolphins QB Chad Pennington completed a 2-yard TD pass to TE Anthony Fasano.  The Bills responded with kicker Rian Lindell getting a 19-yard field goal.  In the second quarter, Buffalo took the lead as Lindell got a 43-yard and a 47-yard field goal...} & \texttt{Which team allowed the most first half points?} & \texttt{Dolphins} \\
  \midrule
  MRQA HotpotQA & \texttt{[PAR] [TLE] John M. Brown [SEP] John Mifflin Brown (September 8, 1817 – March 16, 1893) was a bishop in the African Methodist Episcopal (AME) church.  He was a leader in the underground railroad.  He helped open a number of churches and schools, including the Payne Institute which became Allen University in Columbia, South Carolina and Paul Quinn College in Waco, Texas.  He was also an early principal of Union Seminary which became Wilberforce University [PAR] [TLE] Waco, Texas [SEP] Waco ( ) is a city which is the county seat of McLennan County, Texas, United States.  It is situated along the Brazos River and I-35, halfway between Dallas and Austin.  The city had a 2010 population of 124,805, making it the 22nd-most populous city in the state.  The US Census 2016 population estimate is 134,432 The Waco Metropolitan Statistical Area consists of McLennan and Falls Counties, which had a 2010 population of 234,906.  Falls County was added to the Waco MSA in 2013.  The US Census 2016 population estimate for the Waco MSA is 265,207.} & \texttt{What city is the home to Paul Quinn College and sets on the Brazos River between Dallas and Austin?} & \texttt{Waco, Texas} \\
  \midrule
  QAMR & \texttt{An additional problem to face the empire came as a result of the involvement of Emperor Maurice -LRB- r. 582 -- 602 -RRB- in Persian politics when he intervened in a succession dispute . This led to a period of peace , but when Maurice was overthrown , the Persians invaded and during the reign of Emperor Heraclius -LRB- r. 610 -- 641 -RRB- controlled large chunks of the empire , including Egypt , Syria , and Anatolia until Heraclius ' successful counterattack . In 628 the empire secured a peace treaty and recovered all of its lost territories .} & \texttt{Whose politics did the empire get involved with?} & \texttt{Persian} \\
\bottomrule
\end{tabular}
\caption{Example passages, questions, and answers from the existing \humanconstructedbenchmarks{} we study.}
\label{tab:existing_human_constructed_benchmark_examples}
\end{table}

\clearpage
\newpage

\subsection{Examples From Existing Cloze Benchmarks}\label{appendix:cloze_examples}

Table~\ref{tab:existing_cloze_benchmark_examples} shows examples from the existing cloze benchmarks we study.

\begin{table}[!h]
\centering
\footnotesize
\begin{tabular}{L{2.75cm}p{7.25cm}p{2.8cm}L{1.25cm}}
\toprule
{\bf Benchmark} & {\bf Passage} (some parts shortened with \texttt{...}) & {\bf Question} & {\bf Answer} \\
 \midrule
  Children's Book Test \newline (Common Nouns) & \texttt{... Lady Latifa argued and urged her wishes , but in vain ; the prince was not to be moved . Then she called to the cupbearers for new wine , for she thought that when his head was hot with it he might consent to stay . The pure , clear wine was brought ; she filled a cup and gave to him . He said : ' O most enchanting sweetheart ! it is the rule for the host to drink first and then the guest . '} & \texttt{So to make him lose his head , she drained the XXXXX ; then filled it again and gave him .} & \texttt{cup} \\
  \midrule
  Children's Book Test \newline (Named Entities) & \texttt{... At last , however , the Sunball became aware how sad Letiko was . ... Then he sent them away , and called two hares to him , and said : ` Will you take Letiko home to her mother ? ' ` Yes , why not ? ' ` What will you eat and drink if you should become hungry and thirsty by the way ? ' ` We will eat grass and drink from streamlets . ' ` Then take her , and bring her home . '} & \texttt{Then the hares set out , taking XXXXX with them , and because it was a long way to her home they became hungry by the way .} & \texttt{Letiko} \\
  \midrule
  LAMBADA & \texttt{sorry 's not going to win me my game tomorrow . my racket is . i ca n't believe i let you take it out of here in the first place ! '' `` but , dad , i 'm sure you made mistakes when you were a hippie teenager ! '' `` and i paid for them !} & \texttt{like you 're going to pay for my} & \texttt{racket} \\
  \midrule
  CNN & \texttt{( @entity0 ) you 'll see some familiar faces in the @entity1 . @entity2 beat @entity3 66 - 52 on sunday , giving @entity4 ' coach @entity5 his 12th trip to the semifinals of the @entity6 men 's basketball tournament . @entity7 and @entity8 each scored 16 to help @entity2 win the @entity9 . @entity3 , led by 16 points from @entity10 , was hoping to earn its first trip to the @entity1 . here 's how the @entity1 , to be played in @entity11 , has shaped up : next saturday , @entity2 will face @entity12 in the first semifinal . in the next game , top seed @entity13 will battle @entity14 . ...} & \texttt{the @entity1 matchups : @placeholder vs. @entity12 and @entity13 vs. @entity14} & \texttt{@entity2} \\
  \midrule
  ReCoRD & \texttt{Secretary of State Hillary Clinton on Monday tried to douse a political firestorm over the deadly assault on a U.S. diplomatic mission in Libya, saying she's responsible for the security of American diplomatic outposts. "I take responsibility," Clinton told CNN in an interview while on a visit to Peru. "I'm in charge of the State Department's 60,000-plus people all over the world, 275 posts. The president and the vice president wouldn't be knowledgeable about specific decisions that are made by security professionals. They're the ones who weigh all of the threats and the risks and the needs and make a considered decision."\newline
@highlight\newline
"What I want to avoid is some kind of political gotcha or blame game,"  Clinton says\newline
@highlight\newline
"I take this very personally," she says\newline
@highlight\newline
Diplomats need security but "can't hang out behind walls," she adds} & \texttt{Clinton also described a desperate scene in the @placeholder during the hours of the attack, as staff tried to find out what had happened.} & \texttt{State Department} \\
\bottomrule
\end{tabular}
\caption{Example passages, questions, and answers from the existing cloze benchmarks we study.}
\label{tab:existing_cloze_benchmark_examples}
\end{table}

\clearpage
\newpage

\subsection{Examples From Existing Synthetic Benchmarks}\label{appendix:synthetic_examples}

Table~\ref{tab:existing_synthetic_benchmark_examples} shows examples from the existing synthetic benchmarks we study. The contents of this table are reproduced from \citet{Weston2016TowardsAQ}.

\begin{table}[!h]
\centering
\footnotesize
\begin{tabular}{L{4cm}p{5cm}p{2.8cm}L{1.25cm}}
\toprule
{\bf Benchmark} & {\bf Passage} & {\bf Question} & {\bf Answer} \\
 \midrule
  bAbI Task 1 \newline (Single Supporting Fact) & \texttt{Mary went to the bathroom. John moved to the hallway. Mary travelled to the office.} & \texttt{Where is Mary?} & \texttt{office} \\
  \midrule
  bAbI Task 2 \newline (Two Supporting Facts) & \texttt{John is in the playground. John picked up the football. Bob went to the kitchen.} & \texttt{Where is the football?} & \texttt{playground} \\
 \midrule
  bAbI Task 3 \newline (Three Supporting Facts) & \texttt{John picked up the apple. John went to the office. John went to the kitchen. John dropped the apple.} & \texttt{Where was the apple before the kitchen?} & \texttt{office} \\
  \midrule
  bAbI Task 4 \newline (Two Argument Relations) & \texttt{The office is north of the bedroom. The bedroom is north of the bathroom. The kitchen is west of the garden.} & \texttt{What is north of the bedroom?} & \texttt{office} \\
  \midrule
  bAbI Task 5 \newline (Three Argument Relations) & \texttt{Mary gave the cake to Fred. Fred gave the cake to Bill. Jeff was given the milk by Bill.} & \texttt{Who did Fred give the cake to?} & \texttt{Bill} \\
  \midrule
  bAbI Task 11 \newline (Basic Coreference) & \texttt{Daniel was in the kitchen. Then he went to the studio. Sandra was in the office.} & \texttt{Where is Daniel?} & \texttt{studio} \\
  \midrule
  bAbI Task 12 \newline (Conjunction) & \texttt{Mary and Jeff went to the kitchen. Then Jeff went to the park.} & \texttt{Where is Jeff?} & \texttt{park} \\
  \midrule
  bAbI Task 13 \newline (Compound Coreference) & \texttt{Daniel and Sandra journeyed to the office. Then they went to the garden. Sandra and John travelled to the kitchen. After that they moved to the hallway.} & \texttt{Where is Daniel?} & \texttt{garden} \\
  \midrule
  bAbI Task 14 \newline (Time Reasoning) & \texttt{In the afternoon Julie went to the park. Yesterday Julie was at school. Julie went to the cinema this evening.} & \texttt{Where did Julie go after the park?} & \texttt{cinema} \\
  \midrule
  bAbI Task 15 \newline (Basic Deduction) & \texttt{Sheep are afraid of wolves. Cats are afraid of dogs. Mice are afraid of cats. Gertrude is a sheep.} & \texttt{What is Gertrude afraid of?} & \texttt{wolves} \\
  \midrule
  bAbI Task 16 \newline (Basic Induction) & \texttt{Lily is a swan. Lily is white. Bernhard is green. Greg is a swan.} & \texttt{What color is Greg?} & \texttt{white} \\
\bottomrule
\end{tabular}
\caption{Example passages, questions, and answers from the existing synthetic benchmarks we study.}
\label{tab:existing_synthetic_benchmark_examples}
\end{table}

\newpage

\section{Full Results on Existing Benchmarks}\label{appendix:existing_full_results}

\subsection{Full Results on Existing Human-Constructed Benchmarks}\label{appendix:existing_human_constructed_benchmarks_full_results}

Table~\ref{tab:human_constructed_results0} and Table~\ref{tab:human_constructed_results1} show the performance of each modeling approach on each existing human-constructed benchmark.

\begin{table}[!h]
\footnotesize
\rowcolors{1}{}{lightgray}
\begin{tabular*}{\textwidth}{@{}>{\columncolor{white}[0pt][\tabcolsep]}p{6cm}R{2.85cm}R{2.85cm}R{2.85cm}}
\toprule
{} &  MRQA NewsQA &  MRQA NaturalQuestions &  MRQA DROP \\

\midrule
RaSoR                                     &        44.68 &                  60.02 &      51.30 \\
BiDAF                                     &        43.49 &                  58.43 &      51.36 \\
DocumentReader                            &        46.30 &                  59.08 &      54.96 \\
DocumentReader (no external features)     &        46.32 &                  59.39 &      54.69 \\
BiDAF++                                   &        46.53 &                  60.23 &      55.16 \\
MnemonicReader                            &        48.43 &                  61.53 &      57.02 \\
MnemonicReader (no external features)     &        48.01 &                  61.80 &      57.35 \\
QANet                                     &        47.03 &                  61.74 &      54.56 \\
FusionNet                                 &        49.00 &                  59.62 &      57.82 \\
FusionNet (no external features)          &        48.88 &                  59.54 &      57.95 \\
BERT (base, uncased)                      &        52.61 &                  67.16 &      52.63 \\
BERT (large, uncased)                     &        54.99 &                  69.38 &      61.54 \\
BERT (large, uncased, whole-word masking) &        57.86 &                  71.67 &      71.66 \\
ALBERT (base, V1)                         &        53.25 &                  67.37 &      61.21 \\
ALBERT (xxlarge, V1)                      &        61.16 &                  72.95 &      78.64 \\
RoBERTa (base)                            &        56.62 &                  68.28 &      64.54 \\
RoBERTa (large)                           &        59.14 &                  72.06 &      74.12 \\
ELECTRA (base)                            &        57.60 &                  70.23 &      69.00 \\
SpanBERT (base)                           &        55.60 &                  69.51 &      63.74 \\
SpanBERT (large)                          &        59.09 &                  72.13 &      75.05 \\
\bottomrule
\end{tabular*}
\caption{Performance of modeling approaches when evaluated on MRQA NewsQA, MRQA NaturalQuestions and MRQA DROP.}
\label{tab:human_constructed_results0}
\end{table}
\begin{table}[!h]
\footnotesize
\rowcolors{1}{}{lightgray}
\begin{tabular*}{\textwidth}{@{}>{\columncolor{white}[0pt][\tabcolsep]}p{6cm}R{4.48cm}R{4.48cm}}
\toprule
{} &  MRQA HotpotQA &   QAMR \\

\midrule
RaSoR                                     &          51.35 &  51.56 \\
BiDAF                                     &          50.94 &  51.84 \\
DocumentReader                            &          52.74 &  56.00 \\
DocumentReader (no external features)     &          52.18 &  54.14 \\
BiDAF++                                   &          53.86 &  54.69 \\
MnemonicReader                            &          56.13 &  58.07 \\
MnemonicReader (no external features)     &          55.60 &  56.92 \\
QANet                                     &          54.16 &  53.31 \\
FusionNet                                 &          57.69 &  59.14 \\
FusionNet (no external features)          &          57.38 &  56.91 \\
BERT (base, uncased)                      &          59.53 &  64.36 \\
BERT (large, uncased)                     &          61.63 &  67.51 \\
BERT (large, uncased, whole-word masking) &          65.02 &  71.03 \\
ALBERT (base, V1)                         &          61.65 &  66.30 \\
ALBERT (xxlarge, V1)                      &          68.17 &  74.15 \\
RoBERTa (base)                            &          61.19 &  67.16 \\
RoBERTa (large)                           &          64.58 &  71.44 \\
ELECTRA (base)                            &          62.58 &  68.16 \\
SpanBERT (base)                           &          63.89 &  68.70 \\
SpanBERT (large)                          &          66.60 &  71.46 \\
\bottomrule
\end{tabular*}
\caption{Performance of modeling approaches when evaluated on MRQA HotpotQA and QAMR.}
\label{tab:human_constructed_results1}
\end{table}

\newpage

\subsection{Full Results on Existing Cloze Benchmarks}\label{appendix:existing_cloze_benchmarks_full_results}

Table~\ref{tab:cloze_results0} and Table~\ref{tab:cloze_results1} show the performance of each modeling approach on each existing cloze benchmark.

\begin{table}[!h]
\footnotesize
\rowcolors{1}{}{lightgray}
\begin{tabular*}{\textwidth}{@{}>{\columncolor{white}[0pt][\tabcolsep]}p{6cm}R{2.85cm}R{2.85cm}R{2.85cm}}
\toprule
{} &  CBT (CN) &  CBT (NE) &  LAMBADA \\

\midrule
RaSoR                                     &     53.00 &     69.85 &    71.95 \\
BiDAF                                     &     52.45 &     72.75 &    70.29 \\
DocumentReader                            &     56.55 &     73.85 &    74.42 \\
DocumentReader (no external features)     &     57.15 &     74.60 &    74.08 \\
BiDAF++                                   &     58.40 &     77.15 &    71.95 \\
MnemonicReader                            &     61.45 &     78.80 &    74.57 \\
MnemonicReader (no external features)     &     61.20 &     77.90 &    74.55 \\
QANet                                     &     57.65 &     76.95 &    74.89 \\
FusionNet                                 &     65.05 &     80.25 &    76.83 \\
FusionNet (no external features)          &     64.85 &     79.85 &    76.92 \\
BERT (base, uncased)                      &     72.40 &     82.45 &    84.13 \\
BERT (large, uncased)                     &     76.65 &     84.55 &    86.83 \\
BERT (large, uncased, whole-word masking) &     79.90 &     86.90 &    91.23 \\
ALBERT (base, V1)                         &     70.75 &     82.70 &    82.14 \\
ALBERT (xxlarge, V1)                      &     86.90 &     90.70 &    94.53 \\
RoBERTa (base)                            &     75.70 &     84.90 &    86.48 \\
RoBERTa (large)                           &     82.45 &     88.60 &    92.27 \\
ELECTRA (base)                            &     74.20 &     84.40 &    86.40 \\
SpanBERT (base)                           &     75.90 &     85.50 &    87.10 \\
SpanBERT (large)                          &     80.75 &     88.80 &    91.65 \\
\bottomrule
\end{tabular*}
\caption{Performance of modeling approaches when evaluated on CBT (CN), CBT (NE) and LAMBADA.}
\label{tab:cloze_results0}
\end{table}
\begin{table}[!h]
\footnotesize
\rowcolors{1}{}{lightgray}
\begin{tabular*}{\textwidth}{@{}>{\columncolor{white}[0pt][\tabcolsep]}p{6cm}R{4.48cm}R{4.48cm}}
\toprule
{} &  CNN (100K Examples) &  ReCoRD \\

\midrule
RaSoR                                     &                74.59 &   32.97 \\
BiDAF                                     &                75.59 &   30.88 \\
DocumentReader                            &                72.66 &   29.97 \\
DocumentReader (no external features)     &                72.38 &   29.52 \\
BiDAF++                                   &                79.20 &   34.93 \\
MnemonicReader                            &                79.46 &   39.01 \\
MnemonicReader (no external features)     &                78.95 &   37.87 \\
QANet                                     &                79.00 &   33.46 \\
FusionNet                                 &                79.05 &   30.89 \\
FusionNet (no external features)          &                78.80 &   28.91 \\
BERT (base, uncased)                      &                79.74 &   58.45 \\
BERT (large, uncased)                     &                82.54 &   67.18 \\
BERT (large, uncased, whole-word masking) &                82.72 &   72.85 \\
ALBERT (base, V1)                         &                79.33 &   56.54 \\
ALBERT (xxlarge, V1)                      &                86.03 &   81.87 \\
RoBERTa (base)                            &                82.26 &   68.88 \\
RoBERTa (large)                           &                86.77 &   77.63 \\
ELECTRA (base)                            &                82.08 &   69.61 \\
SpanBERT (base)                           &                83.31 &   69.23 \\
SpanBERT (large)                          &                84.81 &   77.72 \\
\bottomrule
\end{tabular*}
\caption{Performance of modeling approaches when evaluated on CNN (100K Examples) and ReCoRD.}
\label{tab:cloze_results1}
\end{table}

\newpage

\subsection{Full Results on Existing Synthetic Benchmarks}\label{appendix:existing_synthetic_benchmarks_full_results}

Table~\ref{tab:babi_results0} and Table~\ref{tab:babi_results1} and Table~\ref{tab:babi_results2} show the performance of each modeling approach on each existing of the bAbI tasks (900 training examples).

\begin{table}[!h]
\footnotesize
\rowcolors{1}{}{lightgray}
\begin{tabular*}{\textwidth}{@{}>{\columncolor{white}[0pt][\tabcolsep]}p{6.1cm}R{2cm}R{2cm}R{2cm}R{2cm}}
\toprule
{} &  bAbI QA \#1 &  bAbI QA \#2 &  bAbI QA \#3 &  bAbI QA \#4 \\

\midrule
RaSoR                                     &       100.0 &        60.0 &        71.0 &        81.0 \\
BiDAF                                     &       100.0 &        42.0 &        53.0 &        83.0 \\
DocumentReader                            &       100.0 &        63.0 &        70.0 &       100.0 \\
DocumentReader (no external features)     &       100.0 &        76.0 &        93.0 &       100.0 \\
BiDAF++                                   &       100.0 &       100.0 &       100.0 &        78.0 \\
MnemonicReader                            &       100.0 &        44.0 &        71.0 &       100.0 \\
MnemonicReader (no external features)     &       100.0 &       100.0 &        74.0 &       100.0 \\
QANet                                     &       100.0 &        42.0 &        39.0 &        85.0 \\
FusionNet                                 &       100.0 &        84.0 &        77.0 &       100.0 \\
FusionNet (no external features)          &       100.0 &       100.0 &        70.0 &       100.0 \\
BERT (base, uncased)                      &       100.0 &        80.0 &        49.0 &        81.0 \\
BERT (large, uncased)                     &       100.0 &        63.0 &        63.0 &        79.0 \\
BERT (large, uncased, whole-word masking) &       100.0 &        98.0 &        98.0 &        91.0 \\
ALBERT (base, V1)                         &       100.0 &        86.0 &        85.0 &        85.0 \\
ALBERT (xxlarge, V1)                      &       100.0 &       100.0 &       100.0 &       100.0 \\
RoBERTa (base)                            &       100.0 &        73.0 &        54.0 &        64.0 \\
RoBERTa (large)                           &       100.0 &        39.0 &        53.0 &        87.0 \\
ELECTRA (base)                            &       100.0 &        86.0 &        64.0 &       100.0 \\
SpanBERT (base)                           &        57.0 &         9.0 &        22.0 &        60.0 \\
SpanBERT (large)                          &        61.0 &        38.0 &         9.0 &        60.0 \\
\bottomrule
\end{tabular*}
\caption{Performance of modeling approaches when evaluated on bAbI QA \#1, bAbI QA \#2, bAbI QA \#3 and bAbI QA \#4.}
\label{tab:babi_results0}
\end{table}
\begin{table}[!h]
\footnotesize
\rowcolors{1}{}{lightgray}
\begin{tabular*}{\textwidth}{@{}>{\columncolor{white}[0pt][\tabcolsep]}p{6.1cm}R{2cm}R{2cm}R{2cm}R{2cm}}
\toprule
{} &  bAbI QA \#5 &  bAbI QA \#11 &  bAbI QA \#12 &  bAbI QA \#13 \\

\midrule
RaSoR                                     &        98.0 &       100.00 &        100.0 &        100.0 \\
BiDAF                                     &        95.0 &        78.00 &        100.0 &         95.0 \\
DocumentReader                            &        96.0 &       100.00 &        100.0 &        100.0 \\
DocumentReader (no external features)     &        97.0 &       100.00 &        100.0 &        100.0 \\
BiDAF++                                   &        96.0 &       100.00 &        100.0 &         95.0 \\
MnemonicReader                            &        95.0 &       100.00 &        100.0 &         95.0 \\
MnemonicReader (no external features)     &        95.0 &       100.00 &        100.0 &        100.0 \\
QANet                                     &        95.0 &       100.00 &        100.0 &         95.0 \\
FusionNet                                 &        98.0 &       100.00 &        100.0 &        100.0 \\
FusionNet (no external features)          &        98.0 &       100.00 &        100.0 &        100.0 \\
BERT (base, uncased)                      &        95.0 &       100.00 &        100.0 &         97.0 \\
BERT (large, uncased)                     &        95.0 &       100.00 &        100.0 &        100.0 \\
BERT (large, uncased, whole-word masking) &        96.0 &       100.00 &        100.0 &        100.0 \\
ALBERT (base, V1)                         &        95.0 &       100.00 &        100.0 &        100.0 \\
ALBERT (xxlarge, V1)                      &        99.0 &       100.00 &        100.0 &        100.0 \\
RoBERTa (base)                            &        95.0 &        98.99 &         89.0 &         95.0 \\
RoBERTa (large)                           &        98.0 &       100.00 &        100.0 &         95.0 \\
ELECTRA (base)                            &        95.0 &       100.00 &        100.0 &         97.0 \\
SpanBERT (base)                           &        36.0 &        74.75 &         75.0 &         95.0 \\
SpanBERT (large)                          &        43.0 &        81.82 &         77.0 &         95.0 \\
\bottomrule
\end{tabular*}
\caption{Performance of modeling approaches when evaluated on bAbI QA \#5, bAbI QA \#11, bAbI QA \#12 and bAbI QA \#13.}
\label{tab:babi_results1}
\end{table}
\begin{table}[!h]
\footnotesize
\rowcolors{1}{}{lightgray}
\begin{tabular*}{\textwidth}{@{}>{\columncolor{white}[0pt][\tabcolsep]}p{6cm}R{2.85cm}R{2.85cm}R{2.85cm}}
\toprule
{} &  bAbI QA \#14 &  bAbI QA \#15 &  bAbI QA \#16 \\

\midrule
RaSoR                                     &         97.0 &        73.00 &         64.0 \\
BiDAF                                     &         95.0 &        66.00 &         61.0 \\
DocumentReader                            &         96.0 &        68.00 &         63.0 \\
DocumentReader (no external features)     &         99.0 &        68.00 &         64.0 \\
BiDAF++                                   &         92.0 &        65.00 &         61.0 \\
MnemonicReader                            &         99.0 &        63.00 &         65.0 \\
MnemonicReader (no external features)     &         99.0 &        67.00 &         65.0 \\
QANet                                     &         62.0 &        64.00 &         58.0 \\
FusionNet                                 &        100.0 &        69.00 &         64.0 \\
FusionNet (no external features)          &         99.0 &       100.00 &         64.0 \\
BERT (base, uncased)                      &         84.0 &        60.56 &         50.0 \\
BERT (large, uncased)                     &         88.0 &        56.34 &         52.0 \\
BERT (large, uncased, whole-word masking) &         96.0 &       100.00 &         62.0 \\
ALBERT (base, V1)                         &         78.0 &        60.56 &         80.0 \\
ALBERT (xxlarge, V1)                      &        100.0 &       100.00 &        100.0 \\
RoBERTa (base)                            &         81.0 &        61.97 &         47.0 \\
RoBERTa (large)                           &         77.0 &       100.00 &         44.0 \\
ELECTRA (base)                            &         87.0 &       100.00 &         47.0 \\
SpanBERT (base)                           &         37.0 &        46.48 &         36.0 \\
SpanBERT (large)                          &         37.0 &        59.15 &         49.0 \\
\bottomrule
\end{tabular*}
\caption{Performance of modeling approaches when evaluated on bAbI QA \#14, bAbI QA \#15 and bAbI QA \#16.}
\label{tab:babi_results2}
\end{table}

\begin{figure*}[!h]
  \centering
  \includegraphics[width=\textwidth]{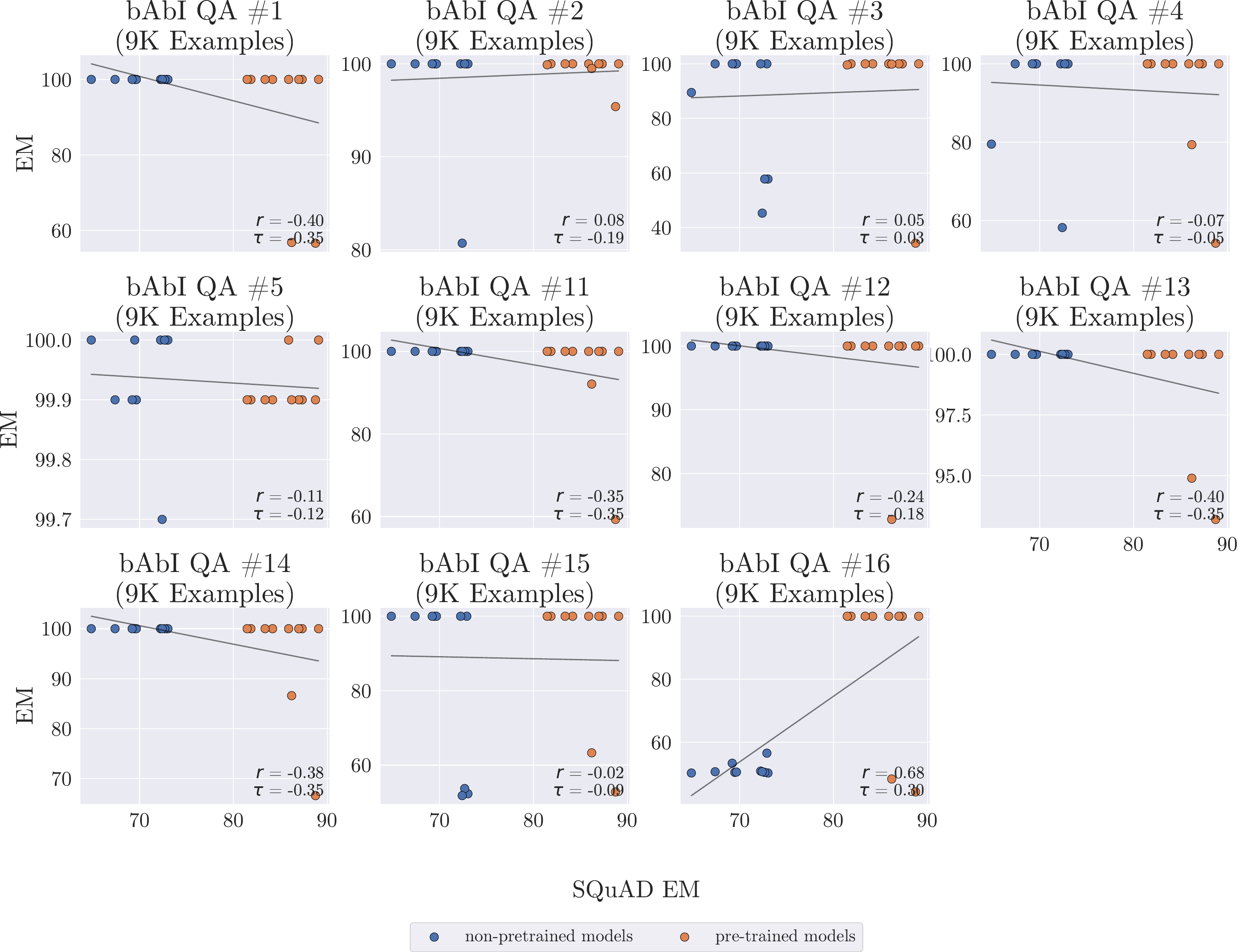}
  \caption{Many modeling approaches perform perfectly on bAbI tasks when training on \numprint{9000} examples, limiting their ability to recapitulate historical modeling progress on SQuAD.}
  \label{fig:squad_vs_synthetic_10k}
\end{figure*}

Figure~\ref{fig:squad_vs_synthetic_10k} shows how well the bAbI tasks (9000) training examples concur with SQuAD.

Table~\ref{tab:babi_9k_results0} and Table~\ref{tab:babi_9k_results1} and Table~\ref{tab:babi_9k_results2} show the performance of each modeling approach on each existing of the bAbI tasks (9000 training examples).

\begin{table}[!h]
\footnotesize
\rowcolors{1}{}{lightgray}
\begin{tabular*}{\textwidth}{@{}>{\columncolor{white}[0pt][\tabcolsep]}p{6.1cm}R{2cm}R{2cm}R{2cm}R{2cm}}
\toprule
{} &  bAbI QA \#1 (9K) &  bAbI QA \#2 (9K) &  bAbI QA \#3 (9K) &  bAbI QA \#4 (9K) \\

\midrule
RaSoR                                     &                    100.00 &                     100.0 &                      89.5 &                     79.50 \\
BiDAF                                     &                    100.00 &                     100.0 &                     100.0 &                    100.00 \\
DocumentReader                            &                    100.00 &                     100.0 &                     100.0 &                    100.00 \\
DocumentReader (no external features)     &                    100.00 &                     100.0 &                     100.0 &                    100.00 \\
BiDAF++                                   &                    100.00 &                     100.0 &                     100.0 &                    100.00 \\
MnemonicReader                            &                    100.00 &                     100.0 &                      57.8 &                    100.00 \\
MnemonicReader (no external features)     &                    100.00 &                     100.0 &                      57.8 &                    100.00 \\
QANet                                     &                    100.00 &                      80.7 &                      45.3 &                     58.20 \\
FusionNet                                 &                    100.00 &                     100.0 &                     100.0 &                    100.00 \\
FusionNet (no external features)          &                    100.00 &                     100.0 &                     100.0 &                    100.00 \\
BERT (base, uncased)                      &                    100.00 &                      99.9 &                      99.6 &                    100.00 \\
BERT (large, uncased)                     &                    100.00 &                     100.0 &                     100.0 &                    100.00 \\
BERT (large, uncased, whole-word masking) &                    100.00 &                     100.0 &                     100.0 &                    100.00 \\
ALBERT (base, V1)                         &                    100.00 &                     100.0 &                     100.0 &                    100.00 \\
ALBERT (xxlarge, V1)                      &                    100.00 &                     100.0 &                     100.0 &                    100.00 \\
RoBERTa (base)                            &                    100.00 &                     100.0 &                     100.0 &                    100.00 \\
RoBERTa (large)                           &                    100.00 &                     100.0 &                     100.0 &                    100.00 \\
ELECTRA (base)                            &                    100.00 &                     100.0 &                     100.0 &                    100.00 \\
SpanBERT (base)                           &                     56.77 &                      99.5 &                      99.9 &                     79.37 \\
SpanBERT (large)                          &                     56.57 &                      95.4 &                      34.3 &                     54.21 \\
\bottomrule
\end{tabular*}
\caption{Performance of modeling approaches when evaluated on bAbI QA \#1 (9K Examples), bAbI QA \#2 (9K Examples), bAbI QA \#3 (9K Examples) and bAbI QA \#4 (9K Examples).}
\label{tab:babi_9k_results0}
\end{table}
\begin{table}[!h]
\footnotesize
\rowcolors{1}{}{lightgray}
\begin{tabular*}{\textwidth}{@{}>{\columncolor{white}[0pt][\tabcolsep]}p{6.1cm}R{2cm}R{2cm}R{2cm}R{2cm}}
\toprule
{} &  bAbI QA \#5 (9K) &  bAbI QA \#11 (9K) &  bAbI QA \#12 (9K) &  bAbI QA \#13 (9K) \\

\midrule
RaSoR                                     &                     100.0 &                     100.00 &                      100.0 &                     100.00 \\
BiDAF                                     &                      99.9 &                     100.00 &                      100.0 &                     100.00 \\
DocumentReader                            &                      99.9 &                     100.00 &                      100.0 &                     100.00 \\
DocumentReader (no external features)     &                      99.9 &                     100.00 &                      100.0 &                     100.00 \\
BiDAF++                                   &                     100.0 &                     100.00 &                      100.0 &                     100.00 \\
MnemonicReader                            &                     100.0 &                     100.00 &                      100.0 &                     100.00 \\
MnemonicReader (no external features)     &                     100.0 &                     100.00 &                      100.0 &                     100.00 \\
QANet                                     &                      99.7 &                     100.00 &                      100.0 &                     100.00 \\
FusionNet                                 &                     100.0 &                     100.00 &                      100.0 &                     100.00 \\
FusionNet (no external features)          &                     100.0 &                     100.00 &                      100.0 &                     100.00 \\
BERT (base, uncased)                      &                      99.9 &                     100.00 &                      100.0 &                     100.00 \\
BERT (large, uncased)                     &                      99.9 &                     100.00 &                      100.0 &                     100.00 \\
BERT (large, uncased, whole-word masking) &                      99.9 &                     100.00 &                      100.0 &                     100.00 \\
ALBERT (base, V1)                         &                      99.9 &                     100.00 &                      100.0 &                     100.00 \\
ALBERT (xxlarge, V1)                      &                     100.0 &                     100.00 &                      100.0 &                     100.00 \\
RoBERTa (base)                            &                      99.9 &                     100.00 &                      100.0 &                     100.00 \\
RoBERTa (large)                           &                      99.9 &                     100.00 &                      100.0 &                     100.00 \\
ELECTRA (base)                            &                     100.0 &                     100.00 &                      100.0 &                     100.00 \\
SpanBERT (base)                           &                      99.9 &                      92.08 &                       72.8 &                      94.89 \\
SpanBERT (large)                          &                      99.9 &                      59.32 &                      100.0 &                      93.19 \\
\bottomrule
\end{tabular*}
\caption{Performance of modeling approaches when evaluated on bAbI QA \#5 (9K Examples), bAbI QA \#11 (9K Examples), bAbI QA \#12 (9K Examples) and bAbI QA \#13 (9K Examples).}
\label{tab:babi_9k_results1}
\end{table}
\begin{table}[!h]
\footnotesize
\rowcolors{1}{}{lightgray}
\begin{tabular*}{\textwidth}{@{}>{\columncolor{white}[0pt][\tabcolsep]}p{6cm}R{2.85cm}R{2.85cm}R{2.85cm}}
\toprule
{} &  bAbI QA \#14 (9K) &  bAbI QA \#15 (9K) &  bAbI QA \#16 (9K) \\

\midrule
RaSoR                                     &                      100.0 &                     100.00 &                       50.2 \\
BiDAF                                     &                      100.0 &                     100.00 &                       50.6 \\
DocumentReader                            &                      100.0 &                     100.00 &                       50.5 \\
DocumentReader (no external features)     &                      100.0 &                     100.00 &                       53.3 \\
BiDAF++                                   &                      100.0 &                     100.00 &                       50.4 \\
MnemonicReader                            &                      100.0 &                      52.30 &                       50.2 \\
MnemonicReader (no external features)     &                      100.0 &                      53.70 &                       50.4 \\
QANet                                     &                      100.0 &                      51.80 &                       50.6 \\
FusionNet                                 &                      100.0 &                     100.00 &                       56.5 \\
FusionNet (no external features)          &                      100.0 &                     100.00 &                       50.8 \\
BERT (base, uncased)                      &                      100.0 &                     100.00 &                      100.0 \\
BERT (large, uncased)                     &                      100.0 &                     100.00 &                      100.0 \\
BERT (large, uncased, whole-word masking) &                      100.0 &                     100.00 &                      100.0 \\
ALBERT (base, V1)                         &                      100.0 &                     100.00 &                      100.0 \\
ALBERT (xxlarge, V1)                      &                      100.0 &                     100.00 &                      100.0 \\
RoBERTa (base)                            &                      100.0 &                     100.00 &                      100.0 \\
RoBERTa (large)                           &                      100.0 &                     100.00 &                      100.0 \\
ELECTRA (base)                            &                      100.0 &                     100.00 &                      100.0 \\
SpanBERT (base)                           &                       86.6 &                      63.30 &                       48.3 \\
SpanBERT (large)                          &                       66.6 &                      52.78 &                       44.2 \\
\bottomrule
\end{tabular*}
\caption{Performance of modeling approaches when evaluated on bAbI QA \#14 (9K Examples), bAbI QA \#15 (9K Examples) and bAbI QA \#16 (9K Examples).}
\label{tab:babi_9k_results2}
\end{table}

\clearpage
\newpage

\section{\fuzzypmsyntheticqa{} Construction Details}\label{appendix:synthetic_patternmatching_construction_details}

\begin{figure}[!h]
  \centering \includegraphics[width=0.6\textwidth]{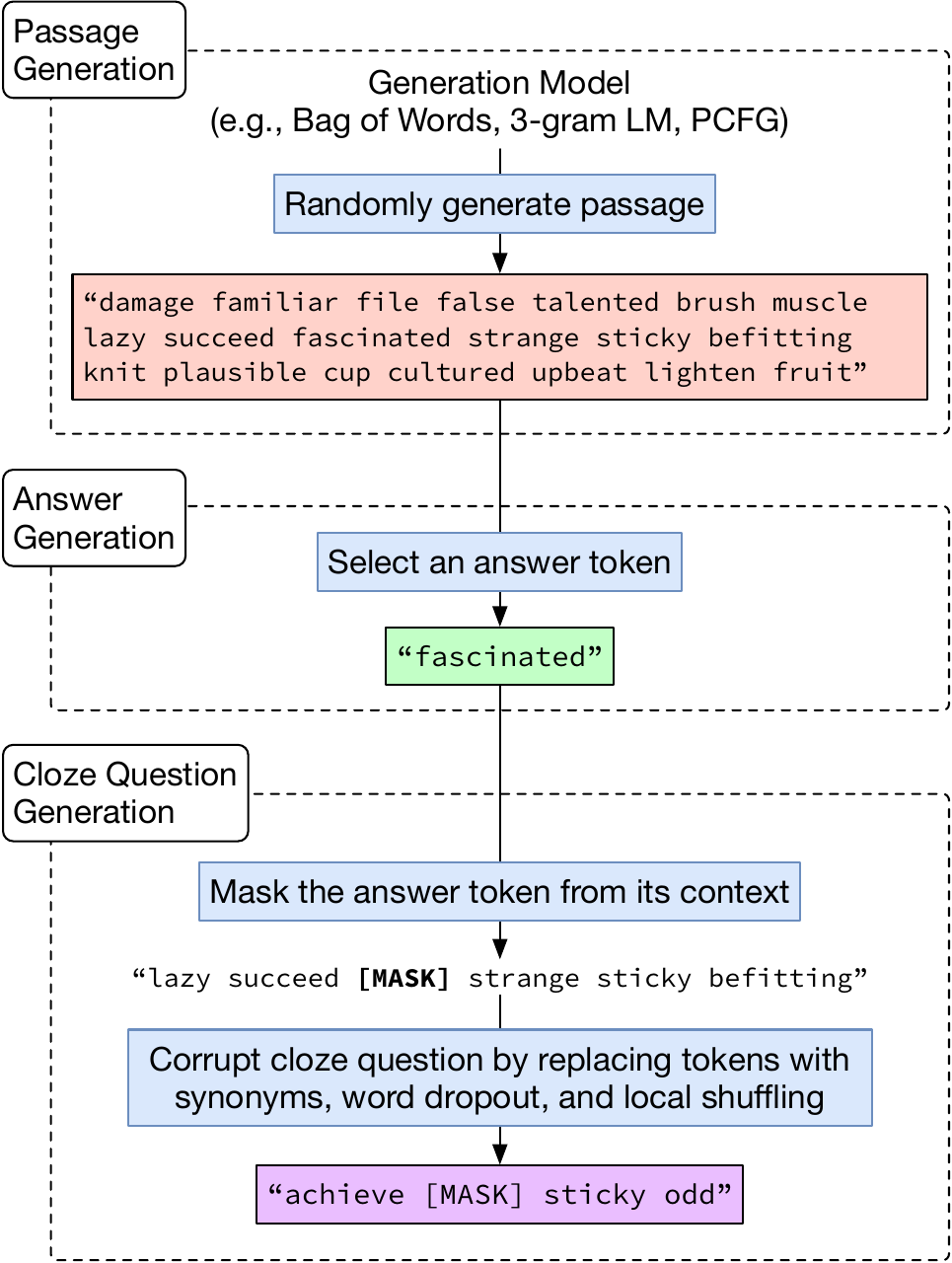}
  \caption{Constructing a \fuzzypmsyntheticqa{} example by generating a \hlc[passage]{passage}, \hlc[answer]{answer}, and \hlc[question]{cloze question}.}
  \label{fig:synthetic_patternmatching_generation}
\end{figure}

Figure~\ref{fig:synthetic_patternmatching_generation} provides an overview of the construction of \fuzzypmsyntheticqa{}.

To efficiently replace tokens with related tokens, we consider each token's \numprint{100} \emph{approximate} nearest neighbors as replacement candidates. In particular, we use Annoy \citep{annoy} to perform the approximate nearest neighboor look-ups. Similarities are derived from the Euclidean distance of normalized vectors between two tokens.

\clearpage
\newpage

\section{Full Results on \fuzzypmsyntheticqa{}}\label{appendix:synthetic_fuzzy_pattern_matching_benchmarks_full_results}

\begin{figure}[!h]
  \centering
  \includegraphics[width=\textwidth]{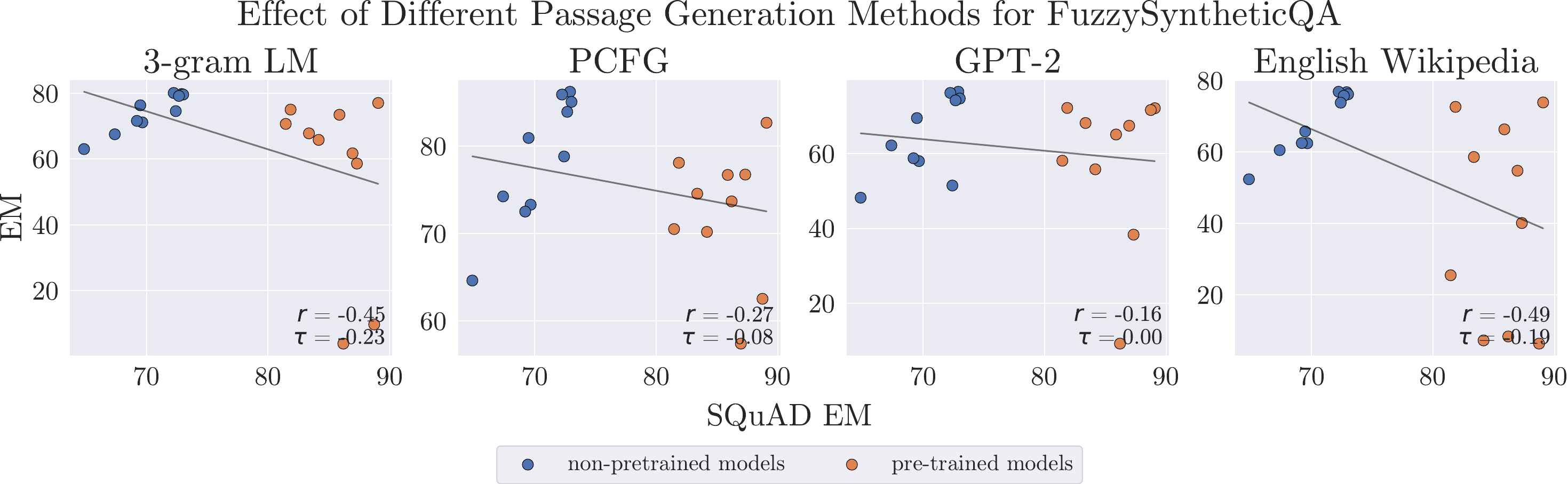}
  \caption{Even with progressively more natural passages, \fuzzypmsyntheticqa continues to have low overall concurrence with SQuAD---this low concurrence is not trivially caused by the lack of natural passages, and simply making our passages more closely resemble natural language will not yield high concurrence.}
  \label{fig:synthetic_patternmatching_generation_ablation}
\end{figure}

Figure~\ref{fig:synthetic_patternmatching_generation_ablation} shows that changing the passage generation method in \fuzzypmsyntheticqa{} has a minimal effect on concurrence. We experiment with generating passages from a 3-gram language model, a probabilistic context-free grammar, a large neural language model (GPT-2 1.5B; \citealp{radford2019language}), and by taking real Wikipedia paragraphs.

The 3-gram language model is trained with maximum likelihood estimation on WikiText-103 \citep{merity2017pointer}. The PCFG is trained with maximum likelihood estimation on the Penn Treebank \citep{Marcus1993PTB}. Lastly, we take GPT-2 1.5B generations from the officially-released output samples (\texttt{\href{https://github.com/openai/gpt-2-output-dataset}{github.com/openai/gpt-2-output-dataset}}; generated with top-k truncated sampling with k = 40).

Table~\ref{tab:synthetic_results0} and Table~\ref{tab:synthetic_results1} show the performance of each modeling approach on each of our constructed synthetic fuzzy pattern-matching benchmarks.

\begin{table}[!h]
\footnotesize
\rowcolors{1}{}{lightgray}
\begin{tabular*}{\textwidth}{@{}>{\columncolor{white}[0pt][\tabcolsep]}p{6cm}R{2.85cm}R{2.85cm}R{2.85cm}}
\toprule
{} &  Synthetic Fuzzy Pattern-Matching &  3-gram LM Synthetic Fuzzy Pattern-Matching &  PCFG Synthetic Fuzzy Pattern-Matching \\

\midrule
RaSoR                                     &                             37.01 &                                       63.00 &                                  64.60 \\
BiDAF                                     &                             38.62 &                                       67.50 &                                  74.23 \\
DocumentReader                            &                             49.32 &                                       71.11 &                                  73.28 \\
DocumentReader (no external features)     &                             49.24 &                                       71.57 &                                  72.49 \\
BiDAF++                                   &                             56.89 &                                       76.30 &                                  80.92 \\
MnemonicReader                            &                             61.50 &                                       79.56 &                                  85.05 \\
MnemonicReader (no external features)     &                             61.24 &                                       79.13 &                                  83.91 \\
QANet                                     &                             59.60 &                                       74.53 &                                  78.80 \\
FusionNet                                 &                             64.71 &                                       79.72 &                                  86.21 \\
FusionNet (no external features)          &                             63.80 &                                       80.05 &                                  85.89 \\
BERT (base, uncased)                      &                              4.51 &                                       70.65 &                                  70.49 \\
BERT (large, uncased)                     &                             40.11 &                                       65.79 &                                  70.17 \\
BERT (large, uncased, whole-word masking) &                              0.70 &                                       58.60 &                                  76.73 \\
ALBERT (base, V1)                         &                             44.28 &                                       75.00 &                                  78.08 \\
ALBERT (xxlarge, V1)                      &                             53.79 &                                       77.01 &                                  82.66 \\
RoBERTa (base)                            &                             44.92 &                                       67.78 &                                  74.54 \\
RoBERTa (large)                           &                              0.49 &                                       61.71 &                                  57.38 \\
ELECTRA (base)                            &                             44.85 &                                       73.42 &                                  76.69 \\
SpanBERT (base)                           &                              0.74 &                                        3.92 &                                  73.66 \\
SpanBERT (large)                          &                              0.40 &                                        9.74 &                                  62.51 \\
\bottomrule
\end{tabular*}
\caption{Performance of modeling approaches when evaluated on Synthetic Fuzzy Pattern-Matching, 3-gram LM Synthetic Fuzzy Pattern-Matching and PCFG Synthetic Fuzzy Pattern-Matching.}
\label{tab:synthetic_results0}
\end{table}
\begin{table}[!h]
\footnotesize
\rowcolors{1}{}{lightgray}
\begin{tabular*}{\textwidth}{@{}>{\columncolor{white}[0pt][\tabcolsep]}p{6cm}R{4.48cm}R{4.48cm}}
\toprule
{} &  GPT-2 Synthetic Fuzzy Pattern-Matching &  English Wikipedia Synthetic Fuzzy Pattern-Matching \\

\midrule
RaSoR                                     &                                   48.20 &                                              52.37 \\
BiDAF                                     &                                   62.16 &                                              60.52 \\
DocumentReader                            &                                   57.97 &                                              62.45 \\
DocumentReader (no external features)     &                                   58.73 &                                              62.50 \\
BiDAF++                                   &                                   69.45 &                                              65.74 \\
MnemonicReader                            &                                   74.67 &                                              76.15 \\
MnemonicReader (no external features)     &                                   74.18 &                                              75.71 \\
QANet                                     &                                   51.45 &                                              73.79 \\
FusionNet                                 &                                   76.48 &                                              76.73 \\
FusionNet (no external features)          &                                   76.17 &                                              76.85 \\
BERT (base, uncased)                      &                                   58.07 &                                              25.52 \\
BERT (large, uncased)                     &                                   55.78 &                                               7.29 \\
BERT (large, uncased, whole-word masking) &                                   38.34 &                                              40.13 \\
ALBERT (base, V1)                         &                                   72.16 &                                              72.62 \\
ALBERT (xxlarge, V1)                      &                                   72.09 &                                              73.86 \\
RoBERTa (base)                            &                                   68.14 &                                              58.60 \\
RoBERTa (large)                           &                                   67.41 &                                              54.76 \\
ELECTRA (base)                            &                                   65.07 &                                              66.33 \\
SpanBERT (base)                           &                                    9.26 &                                               8.40 \\
SpanBERT (large)                          &                                   71.61 &                                               6.40 \\
\bottomrule
\end{tabular*}
\caption{Performance of modeling approaches when evaluated on GPT-2 Synthetic Fuzzy Pattern-Matching and English Wikipedia Synthetic Fuzzy Pattern-Matching.}
\label{tab:synthetic_results1}
\end{table}

\clearpage
\newpage

\section{\wikidatasyntheticqa{} Construction Details}\label{appendix:wikidata_construction_details}

\begin{figure}
  \centering \includegraphics[width=0.6\textwidth]{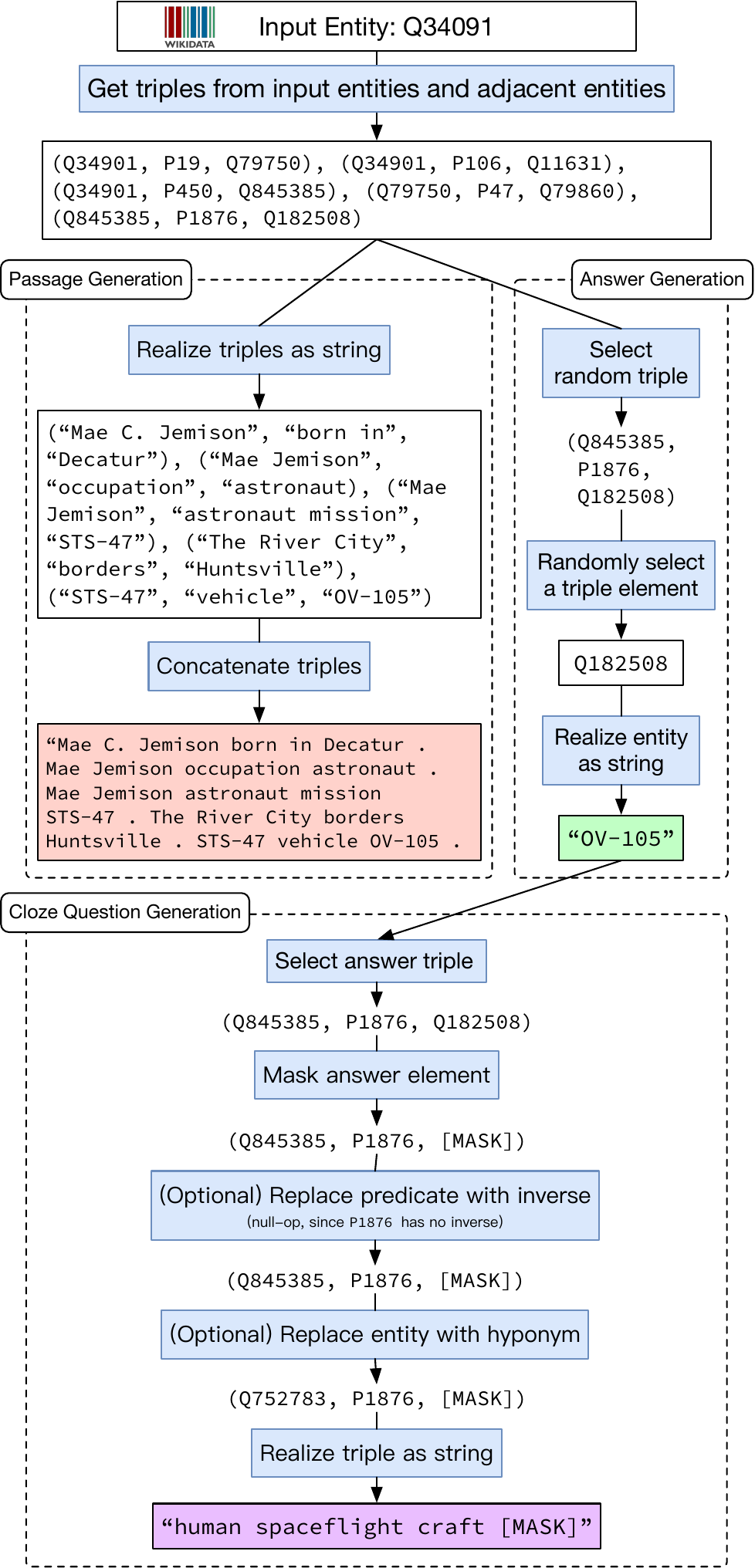}
  \caption{Constructing a \wikidatasyntheticqa{} example by generating a \hlc[passage]{passage}, \hlc[answer]{answer}, and \hlc[question]{cloze question}.}\label{fig:synthetic_wikidata_generation}
\end{figure}

Figure~\ref{fig:synthetic_wikidata_generation} summarizes the data generation procedure for \wikidatasyntheticqa{}.

\paragraph{Inverses of Properties.}

Some of our generated questions use the inverse relationships between two properties. To obtain the inverse relationship for a given property, we first retrieve its list of property constraints by using Wikidata property \texttt{P2302 (property constraint)}. If \texttt{Q21510855 (inverse constraint)} is present, we then retrieve the corresponding property of this inverse relationship. If the inverse constraint is not present, we check the corresponding property of \texttt{P7087 (inverse label item)}, which outputs the item with a label of the inverse relationship of the property.

\paragraph{Entity Hyponyms.}

Some of our generated questions replace entities with their hyponyms. To obtain the hyponyms for a given entity, we retrieve any object entities of the \texttt{P31 (instance of)} and \texttt{P279 (subclass of)} properties.

\clearpage
\newpage

\section{Full Results on \wikidatasyntheticqa{}}\label{appendix:synthetic_wikidata_benchmarks_full_results}

Table~\ref{tab:synthetic_results2} shows the performance of each modeling approach on \wikidatasyntheticqa{}.

\begin{table}[!h]
\footnotesize
\rowcolors{1}{}{lightgray}
\begin{tabular*}{\textwidth}{@{}>{\columncolor{white}[0pt][\tabcolsep]}p{6cm}R{9.35cm}}
\toprule
{} &  Synthetic Wikidata \\

\midrule
RaSoR                                     &               63.67 \\
BiDAF                                     &               68.69 \\
DocumentReader                            &               67.66 \\
DocumentReader (no external features)     &               68.03 \\
BiDAF++                                   &               70.43 \\
MnemonicReader                            &               75.04 \\
MnemonicReader (no external features)     &               74.31 \\
QANet                                     &               73.12 \\
FusionNet                                 &               74.52 \\
FusionNet (no external features)          &               73.90 \\
BERT (base, uncased)                      &               73.68 \\
BERT (large, uncased)                     &               78.01 \\
BERT (large, uncased, whole-word masking) &               81.56 \\
ALBERT (base, V1)                         &               77.23 \\
ALBERT (xxlarge, V1)                      &               86.29 \\
RoBERTa (base)                            &               77.75 \\
RoBERTa (large)                           &               82.79 \\
ELECTRA (base)                            &               76.86 \\
SpanBERT (base)                           &               78.50 \\
SpanBERT (large)                          &               84.26 \\
\bottomrule
\end{tabular*}
\caption{Performance of modeling approaches when evaluated on Synthetic Wikidata.}
\label{tab:synthetic_results2}
\end{table}

\clearpage
\newpage

\section{Full Results on Subsampled SQuAD}\label{appendix:subsampled_squad_full_results}

Table~\ref{tab:squad_results0} and Table~\ref{tab:squad_results1} show the performance of each modeling approach on subsamples of the SQuAD benchmark.

\begin{table}[!h]
\footnotesize
\rowcolors{1}{}{lightgray}
\begin{tabular*}{\textwidth}{@{}>{\columncolor{white}[0pt][\tabcolsep]}p{6cm}R{2.85cm}R{2.85cm}R{2.85cm}}
\toprule
\multirow{5}{*}{} & \multicolumn{3}{c}{SQuAD 1.1}\\ \cmidrule{2-4}
{} &  All &  1K Examples &  10K Examples \\

\midrule
RaSoR                                     &  64.86 &                15.52 &                 49.44 \\
BiDAF                                     &  67.39 &                 7.96 &                 48.54 \\
DocumentReader                            &  69.66 &                34.66 &                 56.42 \\
DocumentReader (no external features)     &  69.21 &                30.69 &                 54.82 \\
BiDAF++                                   &  69.49 &                18.62 &                 57.48 \\
MnemonicReader                            &  73.02 &                30.67 &                 58.91 \\
MnemonicReader (no external features)     &  72.67 &                29.46 &                 57.79 \\
QANet                                     &  72.41 &                 7.18 &                 48.15 \\
FusionNet                                 &  72.90 &                37.52 &                 59.97 \\
FusionNet (no external features)          &  72.24 &                35.55 &                 58.69 \\
BERT (base, uncased)                      &  81.46 &                31.80 &                 70.34 \\
BERT (large, uncased)                     &  84.17 &                49.08 &                 75.47 \\
BERT (large, uncased, whole-word masking) &  87.32 &                69.19 &                 81.78 \\
ALBERT (base, V1)                         &  81.86 &                57.57 &                 74.55 \\
ALBERT (xxlarge, V1)                      &  89.07 &                76.36 &                 86.19 \\
RoBERTa (base)                            &  83.37 &                55.01 &                 77.30 \\
RoBERTa (large)                           &  86.96 &                62.64 &                 82.56 \\
ELECTRA (base)                            &  85.88 &                62.05 &                 78.31 \\
SpanBERT (base)                           &  86.20 &                65.80 &                 80.72 \\
SpanBERT (large)                          &  88.74 &                75.00 &                 85.06 \\
\bottomrule
\end{tabular*}
\caption{Performance of modeling approaches when evaluated on SQuAD, SQuAD (1K Examples) and SQuAD (10K Examples).}
\label{tab:squad_results0}
\end{table}
\begin{table}[!h]
\footnotesize
\rowcolors{1}{}{lightgray}
\begin{tabular*}{\textwidth}{@{}>{\columncolor{white}[0pt][\tabcolsep]}p{6cm}R{2.85cm}R{2.85cm}R{2.85cm}}
\toprule
\multirow{5}{*}{} & \multicolumn{3}{c}{SQuAD 1.1}\\ \cmidrule{2-4}
{} &  20K Examples &  40K Examples &  60K Examples \\

\midrule
RaSoR                                     &                 55.13 &                 60.37 &                 62.95 \\
BiDAF                                     &                 57.29 &                 62.35 &                 65.25 \\
DocumentReader                            &                 61.84 &                 65.45 &                 68.27 \\
DocumentReader (no external features)     &                 59.66 &                 64.47 &                 67.09 \\
BiDAF++                                   &                 62.25 &                 66.42 &                 68.62 \\
MnemonicReader                            &                 64.74 &                 69.09 &                 70.86 \\
MnemonicReader (no external features)     &                 63.71 &                 68.65 &                 70.32 \\
QANet                                     &                 61.02 &                 66.55 &                 69.74 \\
FusionNet                                 &                 64.74 &                 69.14 &                 70.98 \\
FusionNet (no external features)          &                 63.28 &                 67.98 &                 69.93 \\
BERT (base, uncased)                      &                 74.84 &                 78.24 &                 80.05 \\
BERT (large, uncased)                     &                 79.27 &                 81.83 &                 83.25 \\
BERT (large, uncased, whole-word masking) &                 84.47 &                 85.78 &                 86.75 \\
ALBERT (base, V1)                         &                 77.05 &                 79.95 &                 81.02 \\
ALBERT (xxlarge, V1)                      &                 86.91 &                 88.02 &                 88.63 \\
RoBERTa (base)                            &                 79.56 &                 81.62 &                 82.37 \\
RoBERTa (large)                           &                 84.26 &                 86.37 &                 87.18 \\
ELECTRA (base)                            &                 81.75 &                 83.95 &                 85.01 \\
SpanBERT (base)                           &                 82.54 &                 84.17 &                 85.39 \\
SpanBERT (large)                          &                 86.21 &                 87.33 &                 87.82 \\
\bottomrule
\end{tabular*}
\caption{Performance of modeling approaches when evaluated on SQuAD (20K Examples), SQUAD (40K Examples) and SQuAD (60K Examples).}
\label{tab:squad_results1}
\end{table}

\end{appendices}

\end{document}